\documentclass[twoside,11pt]{article}

\usepackage{blindtext}

\usepackage[abbrvbib, preprint]{jmlr2e}

\usepackage{lipsum}
\usepackage{amsfonts,amsmath}
\usepackage{graphicx}
\usepackage{epstopdf}
\usepackage{algorithm,algorithmic}
\usepackage{caption,subcaption}
\usepackage{multirow}
\usepackage{booktabs}
\usepackage{enumitem}
\usepackage{tikz-cd}
\usepackage{mathtools}
\usetikzlibrary{tikzmark, positioning, fit, shapes.misc, shapes.multipart}

\def\R{\mathbb{R}} 
\def\E{\mathbb{E}} 
\def\x{\boldsymbol{x}}
 
\def\v{\boldsymbol{v}} 
\def\r{\boldsymbol{r}}

\def\y{\boldsymbol{y}}

\def\p{\partial} 
\def\d{\mathrm{d}}

\def\Y{\mathcal{Y}}

\DeclareMathOperator{\argmin}{argmin}

\newcommand{\TT}{\mathsf{TT}} 
\newcommand{\TF}{\mathsf{TF}}
\newcommand{\pE}{p^\mathsf{E}} 
\newcommand{\pS}{p^\mathsf{S}}
\newcommand{\A}{\mathsf{A}} 
\newcommand{\B}{\mathsf{B}}
\newcommand{\G}{\mathsf{G}}

\usepackage{lastpage}
\jmlrheading{23}{2022}{1-\pageref{LastPage}}{9/22}{}{22-0000}{Yinuo Ren, Hongli
Zhao, Yuehaw Khoo, and Lexing Ying}

\ShortHeadings{High-dimensional density estimation with tensorizing flow}{Ren,
Zhao, Khoo, and Ying}
\firstpageno{1}

\begin{document}

\title{High-dimensional density estimation with tensorizing flow}

\author{\name Yinuo Ren$^1$  \email yinuoren@stanford.edu \AND \name Hongli
	Zhao$^2$ \email honglizhaobob@uchicago.edu \AND \name Yuehaw Khoo$^2$ \email
	ykhoo@uchicago.edu \AND \name Lexing Ying$^{1,3}$ \email lexing@stanford.edu
	\\
	\addr $^1$Institute for Computational and Mathematical Engineering (ICME),
	Stanford University, Stanford, CA 94305, USA\\
	\addr $^2$Department of Statistics, University of Chicago, Chicago, IL
	60637, USA\\
	\addr $^3$Department of Mathematics, Stanford University, Stanford, CA
94305, USA }

\editor{My editor}

\maketitle

\begin{abstract}
	We propose the tensorizing flow method for estimating high-dimensional
	probability density functions from the observed data. The method is based on
	both tensor-train and flow-based generative modeling. Our
	method first efficiently constructs an approximate density in the
	tensor-train form via solving the tensor cores from a linear system based on
	the kernel density estimators of low-dimensional marginals. We then train a
	continuous-time flow model from this tensor-train density to the observed
	empirical distribution by performing a maximum likelihood estimation. The
	proposed method combines the optimization-less feature of the tensor-train
	with the flexibility of the flow-based generative models. Numerical results
	are included to demonstrate the performance of the proposed method.
\end{abstract}


\section{Introduction}

Density estimation is one of the most important tasks of statistics and plays a
crucial role in statistical inference, machine learning, and data analysis. It
aims to reconstruct the underlying probability density function directly from
observed data. The estimation for high-dimensional probability distributions has
remained a main challenge both theoretically and computationally.

In recent years, deep generative modeling has been developed as a popular method
for approximating high dimensional densities from a large number of
samples~\citep{bond2021deep}. Several influential advances in this area include
the variational autoencoder (VAE)~\citep{kingma2013auto} and generative
adversarial network (GAN)~\citep{goodfellow2014generative}. Among them, one
particular powerful kind of methods is the flow-based generative
models~\citep{dinh2014nice,rezende2015variational}, which construct a
parameterized flow from a normal distribution to the target distribution.
However, confining the source distribution to a normal distribution (or a
mixture of Gaussians) can be quite restrictive, especially when dealing with 
singular or multimodal distributions.

The tensor-train network~\citep{oseledets2011tensor}, also known as the matrix
product state (MPS)~\citep{perez2006matrix} in the physics literature, is a
class of tensor network structures widely used to model high-dimensional
functions, such as the wavefunction of many-body quantum states under certain
correlation decay assumptions~\citep{brandao2015exponential}.  Inspired by its
success in physics, the tensor-train structure has also been applied to many
other contexts. Various methods are proposed for efficiently obtaining the
tensor-train representation of closed-form high-dimensional functions by the
techniques of linear algebra~\citep{oseledets2010tt},
optimization~\citep{savostyanov2011fast} and parallel
computing~\citep{shi2021parallel}. More recently, \citet{hur2022generative}
proposes an optimization-less linear algebra framework for recovering the
tensor-train representation of a density directly from its empirical
distribution. However, tensor-trains can be inflexible when it comes to machine
learning applications. When working with tensor-train, one needs to
pre-determine the ordering of the variables according to their correlations,
which may be difficult in practice. A sub-optimal ordering may result in larger
ranks of the tensor-train and hence higher storage and computational complexity.

\subsection{Contributions}

Motivated by the strengths as well as the limitations of the flow-based
generative models and the tensor-train representation, we propose a new
framework that combines the benefits of both approaches for density estimation.
More specifically,
\begin{itemize}
	\item We first construct a low-rank approximate tensor-train representation
	      directly from samples as the base distribution. In particular, in
	      order to deal with the data sparsity in high-dimensional settings, the
	      required low-order marginals during the construction are estimated by
	      the kernel density estimation.
	\item We then adopt an ODE-based continuous-time flow model that maps the
	      approximate tensor-train distribution to the target distribution. The
	      flow model is parameterized by a neural network and is trained based
	      on the maximum likelihood estimation with both forward and inverse map
	      computed efficiently.
\end{itemize}
Following the work by~\citet{KLZ2022TF}, we refer to this method as the
\emph{tensorizing flow} approach for density estimation.

\subsection{Related works}

\subsubsection*{Deep generative modeling}
One of the earliest deep generative models is the Boltzmann
machine~\citep{hinton1983optimal,hinton2002training}, which is based on a
certain supposed form of the energy function of the probability distribution.
The method of VAE~\citep{kingma2013auto,rezende2014stochastic} considers an
encoding of the observations in a regularized latent space.  Other approaches
include GAN~\citep{goodfellow2014generative}, which consists of a generator and
a discriminator trained jointly as a minimax
game~\citep{schmidhuber2020generative}, as well as the autoregressive likelihood
models~\citep{bengio2000neural, larochelle2011neural, germain2015made}, which
are based on the chain rule of probability.

Flow-based generative modeling~\citep{dinh2014nice,rezende2015variational} is
another deep generative modeling technique based on a sequence of
diffeomorphisms between a known base distribution and a target distribution of
interest. Unlike the autoregressive models and VAEs, the transformation
performed in flow-based models must be invertible and the determinant of its
Jacobian should be computed efficiently~\citep{kobyzev2020normalizing}. Several
widely-adopted architectures are the planar flow~\citep{rezende2015variational},
coupling flow~\citep{dinh2014nice,dinh2016density}, autoregressive
flow~\citep{kingma2016improved,papamakarios2017masked}, 1$\times$1
convolution~\citep{kingma2018glow}, and spline
flow~\citep{durkan2019cubic,durkan2019neural}. Residual networks use residual
connections to build a reversible network, as in
RevNets~\citep{gomez2017reversible}, iRevNets~\citep{jacobsen2018revnet}, and
iResNet~\citep{behrmann2019invertible}.

The idea of residual connections can be generalized to continuous-time or
infinitesimal flow models. One type of the continuous-time flow models is
formulated by the theory of ordinary differential equations (ODEs)
~\citep{chen2018neural,grathwohl2018ffjord,dupont2019augmented}, among which
\citet{zhang2018monge} proposes a continuous-time gradient flow model from the
perspective of optimal transport and fluid dynamics. The other type is based on
diffusion processes and formulated by stochastic differential equations
(SDEs)~\citep{tabak2010density,chen2018continuous,tzen2019neural}.

\subsubsection*{Tensor-train representation}

The tensor-train representation originates from the density-matrix
renormalization group (DMRG)~\citep{white1993density} in physics and plays
important roles in computational
mathematics~\citep{delathauwer2000multilinear,delathauwer21best,grasedyck2010hierarchical}.
It sees successful applications in high-dimensional scientific computing
problems~\citep{dolgov2014computation,kressner2016low,bachmayr2016tensor},
quantum chemistry and molecular physics~\citep{chan2011density,
baiardi2020density}, and signal and image
processing~\citep{cichocki2009nonnegative,wang2018tensor}.  Many methods have
been proposed for constructing the low-rank tensor-train representation of
high-dimensional functions in the scenario in which one is able to evaluate the
function at arbitrary points or with limited number of evaluations, such as
TT-cross~\citep{oseledets2010tt} and DMRG-cross~\citep{savostyanov2011fast},
TT completion~\citep{steinlechner2016riemannian}, and STTA~\citep{kressner2022streaming}.

The tensor-train representation has also been recently applied to generative
modeling, in which scenario one constructs the tensor-train model directly from
samples without the access to the values of the density function.  Several
earliest attempts are based on the optimization-based DMRG
scheme~\citep{han2018unsupervised,bradley2020modeling}, and Riemannian
optimization~\citep{novikov2021tensor}.  In the work
by~\citet{hur2022generative}, an optimizationless method is proposed for
constructing the tensor-train representation directly from samples using a
sketching technique.  Some other works also manage to take advantage of other
structures of tensor networks, including the tree tensor
network~\citep{cheng2019tree,tang2022generative}, and the projected
entangled-pair state (PEPS)~\citep{vieijra2022generative}.

\subsubsection*{Variational inference via tensorizing flow}

Density estimation via the maximum likelihood estimation is closely related to
the variational inference (VI) problem. Variational inference aims to
approximate an unnormalized density with a low-complexity ansatz by solving an
optimization problem over variational parameters~\citep{blei2017variational}.
Early approaches include mean-field VI, coordinate ascend
VI~\citep{bishop2006pattern}, stochastic VI~\citep{hoffman2013stochastic}, and
black box VI~\citep{ranganath2014black}. Recently, deep neural networks have
also been actively applied in this field~\citep{mnih2014neural,miao2016neural}.
In a manuscript~\citep{KLZ2022TF}, tensorizing flow, \emph{i.e.} the combination
of a tensor-based distribution and the neural network, is suggested for
variational inference problems, in which an unnormalized analytic form of the
density is given for constructing the approximate tensor-train representation.
In contrast, our current density estimation task constructs the approximate
density only based on limited given samples, and consequently a different set of
techniques requires developing so as to deal with the challenges therein.

\subsection{Organization}

The paper is organized as follows. In Section~\ref{sec:bg}, we introduce some
preliminaries. Our proposed method is detailed in Section~\ref{sec:alg}. We
demonstrate the advantage of our proposed method via numerical experiments in
Section~\ref{sec:experiments}. Finally we conclude in
Section~\ref{sec:conclusions} with some discussions of our method.

\section{Problem and Background}
\label{sec:bg}

In this section, we introduce the problem setting and the commonly-used
notations in Section~\ref{sec:notation}, the tensor-train representation for
both tensors and general functions in Section~\ref{sec:tt}, and the
continuous-time flow model in Section~\ref{sec:maflow}.

\subsection{Problem setting and notations}
\label{sec:notation}

We work with probability distributions $p(\x)$ defined on $\R^d$, where
$\x=(x_1,\ldots,x_d)$ with $x_i$ as the individual coordinates. Suppose we are
given $N$ independent $d$-dimensional samples $\{\x^{(i)}=
(x_1^{(i)},\ldots,x_d^{(i)})\}_{1\le i\le N}$ drawn from an unknown distribution
with probability density $p^*(\x):\R^d\rightarrow\R$, the problem is to
construct another probability density $p_\theta(\x)$ with parameter $\theta$
that can serve as an approximation to $p^*(\x)$. The approximation
$p_\theta(\x)$ is also expected to be normalized and easy-to-sample.

Let $\pE(\x)$ be the empirical distribution of the samples, \emph{i.e.}
\begin{equation}
	\pE(\x)=\dfrac{1}{N}\sum_{i=1}^N \delta\left(\x-\x^{(i)}\right).
	\label{eq:emp}
\end{equation}
This task is typically formulated via the maximum likelihood estimation, where
the parameter $\theta$ is obtained by
\begin{equation}
	\begin{aligned}
		\theta = \argmin_{\theta} \mathrm{D}_{\text{KL}} \left(p^*(\cdot)\|p_\theta(\cdot)  \right) & =\argmin_{\theta} \E_{\x\sim p^*}\left[-\log p_\theta(\x)\right]         \\
		                                                                                            & \approx \argmin_{\theta} \E_{\x\sim \pE}\left[-\log p_\theta(\x)\right]. 
	\end{aligned}
	\label{eq:MLE}
\end{equation}

In what follows, we often adopt MATLAB notation in order to simplify the
notations. For example, $m:n$ represents $m,\ldots,n$. For a $3$-tensor $\A$,
$\A(:,i,:)$ denotes the $i$-th slice of the 3-dimensional tensor $\A$ along its
second dimension. We also write $1,\ldots,n$ by $[n]$, variables
$x_m,\ldots,x_n$ by $x_{m:n}$, and the corresponding infinitesimal volume $\d
x_m\cdots \d x_n$ by $\d x_{m:n}$. For a distribution $p(\x)$, the marginal
distribution of variables $x_{m:n}$ is denoted by $p(x_{m:n})$. Especially, the
marginal distributions of variables $x_{1:2}, x_{1:3}, \ldots, x_{d-2:d},
x_{d-1:d}$ are denoted by
\begin{equation}
	p_1(x_{1:2}),\; p_2(x_{1:3}),\ldots,p_{d-1}(x_{d-2:d}),\;  p_d(x_{d-1:d}),
	\label{eq:marginals}
\end{equation} 
among which $p_1$ and $p_d$ are 2-marginals and the rest are 3-marginals.

For simplicity, we often assume $p(\x)$ to be sufficiently smooth and
$\text{supp}(p)\subset I^d$ for an interval $I\subset\R$. In the discussions
throughout Sections~\ref{sec:bg} and~\ref{sec:alg}, we assume $I=[-1,1]$ while
the general cases where $I=[a,b]$ or even $I=\R$ can be handled similarly via
appropriate translations and re-scaling.

\subsection{Tensor-train representation}
\label{sec:tt}

In modern machine learning and scientific computing, data are often presented as
tensors.  A $d$-dimensional tensor $\mathsf{F}$ in $\R^{n \times \cdots \times
n}$ is a collection of numbers denoted by $\mathsf{F}(i_1,\ldots,i_d)$ with
$1\le i_1,\ldots,i_d\le n$. It has $n^d$ elements and is generally impractical
to handle due to its exponential computational cost as the dimension $d$ grows.

One way to represent or approximate high-dimensional tensors is to use the
\emph{tensor-train (TT) representation}, \emph{i.e.}
\begin{equation}
	\mathsf{F}(i_1,\ldots,i_d) \approx
	\G_1(i_1,:)\G_2(:,i_2,:)\cdots \G_d(:,i_d),
	\label{eq:discretett}
\end{equation}
where $\G_1\in \R^{n \times r_1},\G_2\in\R^{r_1\times n \times r_2}, \ldots,
\G_d\in\R^{r_{d-1}\times n}$ are the \emph{cores}, and $r_i$ for $1\le i\le d-1$
are the \emph{ranks} of the TT representation.

\begin{figure}[!htbp]
	\begin{subfigure}[b]{\textwidth}
		\centering
		\begin{tikzpicture}
			            
			\node[circle,draw, scale=2,label= {below: $\G_1$}](1) at (-4,0) {};
			\node[circle,draw, scale=2,label= {below: $\G_2$}](2) at (-2,0) {};
			\node[circle,draw, scale=2,label= {below: $\G_3$}](3) at (0,0) {};
			\node[circle,draw, scale=2,label= {below: $\G_{d-1}$}](d-1) at (3,0)
			{}; \node[circle,draw, scale=2,label= {below: $\G_d$}](d) at (5,0)
			{};
			            
			\node[left = 0.3 of 1] {$\mathsf{F}\approx$};
			            
			\node[above = 0.5 of 1] (1a) {$i_1$}; \node[above = 0.5 of 2] (2a)
			{$i_2$}; \node[above = 0.5 of 3] (3a) {$i_3$}; \node[above = 0.5 of
			d-1] (d-1a) {$i_{d-1}$}; \node[above = 0.5 of d] (da) {$i_d$};
			            
			\draw (1)--node[label={[yshift = -3pt]above:$\alpha_1$}]{} (2);
			\draw (2)--node[label={[yshift = -3pt]above:$\alpha_2$}]{}(3); \draw
			(3)--(1,0); \path (3)-- node[auto=false]{$\cdots$} (d-1); \draw
			(2,0)--(d-1); \draw (d-1)--node[label={[yshift =
			-3pt]above:$\alpha_{d-1}$}]{}(d);
			            
			\draw (1) -- (1a); \draw (2) -- (2a); \draw (3) -- (3a); \draw (d-1)
			-- (d-1a); \draw (d) -- (da);
		\end{tikzpicture}
		\caption{Discrete tensor-train representation}
		\label{fig:discretett}
	\end{subfigure}
	    
	\begin{subfigure}[b]{\textwidth}
		\centering
		\begin{tikzpicture}
			\node[circle,draw, scale=2,label= {below: $G_1$}](1) at (-4,0) {};
			\node[circle,draw, scale=2,label= {below: $G_2$}](2) at (-2,0) {};
			\node[circle,draw, scale=2,label= {below: $G_3$}](3) at (0,0) {};
			\node[circle,draw, scale=2,label= {below: $G_{d-1}$}](d-1) at (3,0)
			{}; \node[circle,draw, scale=2,label= {below: $G_d$}](d) at (5,0)
			{};
			            
			\node[left = 0.3 of 1] {$F\approx$};
			            
			\node[above = 0.5 of 1] (1a) {$x_1$}; \node[above = 0.5 of 2] (2a)
			{$x_2$}; \node[above = 0.5 of 3] (3a) {$x_3$}; \node[above = 0.5 of
			d-1] (d-1a) {$x_{d-1}$}; \node[above = 0.5 of d] (da) {$x_d$};
			            
			\draw (1)--node[label={[yshift = -3pt]above:$\alpha_1$}]{}(2); \draw
			(2)--node[label={[yshift = -3pt]above:$\alpha_2$}]{}(3); \draw
			(3)--(1,0); \path (3)-- node[auto=false]{$\cdots$} (d-1); \draw
			(2,0)--(d-1); \draw (d-1)--node[label={[yshift =
			-3pt]above:$\alpha_{d-1}$}]{}(d);
			            
			\draw[dashed] (1) -- (1a); \draw[dashed] (2) -- (2a); \draw[dashed]
			(3) -- (3a); \draw[dashed] (d-1) -- (d-1a); \draw[dashed] (d) --
			(da);
		\end{tikzpicture}
		\caption{Continuous tensor-train representation}
		\label{fig:continuoustt}
	\end{subfigure}
	\caption{The diagrammatic notation of the tensor-train representation: Solid
		lines represent discrete indices while dashed lines represent continuous
		variables. }
	\label{fig:diagram}
\end{figure}
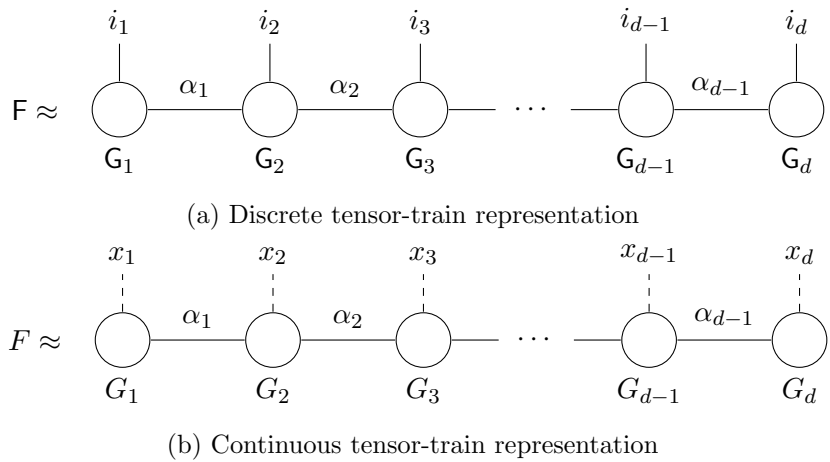

The tensor $\mathsf{F}$ is then represented by the product of a sequence of
corresponding slices of the cores, which is often described in the
\emph{diagrammatic notation} as shown in Figure~\ref{fig:discretett}. We refer
the readers to the discussions by~\citet{penrose1971applications} for the
interpretation of this kind of notations.  When the ranks $\{r_i\}_{1\le i\le
d-1}$ are bounded, TT format features linear cost in $n$ and $d$.

The idea of tensor-train can also be generalized to obtain the low-rank
approximation of high-dimensional functions. The TT representation of a general
$d$-dimensional function $F(\x):I^d\rightarrow \R$ is comprised of a sequence of
$d$ functions $G_1:I\times[r_1]\rightarrow\R$, $G_2:[r_1] \times I \times
[r_2]\rightarrow\R$, $\ldots$, $G_d:[r_{d-1}]\times I\rightarrow\R$, as
\begin{equation}
	F(x_{1:d})\approx
	\sum_{\alpha_1=1}^{r_1}\sum_{\alpha_2=1}^{r_2}\cdots\sum_{\alpha_{d-1}=1}^{r_{d-1}}G_1(x_1,\alpha_1)G_2(\alpha_1,x_2,\alpha_2)\cdots
	G_d(\alpha_{d-1},x_d),
	\label{eq:continuoustt}
\end{equation}
or more compactly
\[
	F(x_{1:d})\approx {G}_1(x_1,:) {G}_2(:,x_2,:)\cdots {G}_d(:,x_d).
\]
The diagrammatic notation of this continuous tensor-train is shown in
Figure~\ref{fig:continuoustt}.

\subsection{Continuous-time flow model}
\label{sec:maflow}

Typically, a flow-based generative model aims to design a pushforward $f:\R^d\rightarrow \R^d$
between a latent easy-to-sample probability density $q_0(\x)$ and a challenging
target probability density $q_1(\x)$ that satisfies
\begin{equation}
	q_1(\x) = q_0\left(f^{-1}(\x)\right)\left|\det\left(\dfrac{\p f^{-1}}{\p \x}\right)\right|.
	\label{eq:jacob}
\end{equation}

A \emph{continuous-time flow model} is based on the perspective that regards $f$ as the
result of a flow that pushes the density $q(\x,t)$ initialized as $q(\x,0) =
q_0(\x)$ over time $t$ with total probability mass being conserved. The
evolution of the density $q(\x,t)$ is characterized by the following
\emph{continuity equation} in the fluid mechanics:
\begin{equation}
	\dfrac{\partial q(\x,t)}{\partial t}+\nabla\cdot\left[q(\x,t)\v(\x)\right]=0,
	\label{eq:continuity}
\end{equation}
where $\v(\x)$ is the velocity field of the flow. Motivated by the linearized
optimal transport, \citet{zhang2018monge} assumes that the flow is irrotational
so that $\nabla\times\v(\x)=0$, and consequently $\v(\x)$ can be written as the
gradient of a potential function $\phi(x)$, \emph{i.e.}
$\v(\x)=\nabla\phi(\x)$~\citep{batchelor2000introduction}.

An equivalent description concerns the trajectory $\x(t)$ that follows the
velocity field $\nabla\phi(\x)$, along which the following two ODEs hold:
\begin{subequations}
	\begin{gather}
		\label{eq:xode}\dfrac{\d \x(t)}{\d t} = \nabla \phi(\x(t)),\\
		\label{eq:pode}\dfrac{\d q(\x(t), t)}{\d t} = -q(\x(t),t) \nabla^2 \phi(\x(t))
	\end{gather}
	\label{eq:ode}
\end{subequations}
where the second equation directly follows from the continuity
equation~\eqref{eq:continuity} and the formula of total derivative $\d/\d t=
\partial/\partial t+\d \x(t)/\d t \cdot
\nabla$~\citep{batchelor2000introduction}. This formulation provides a more
straightforward way to understand the forward map $f$ as the map from $\x(0)$ to
$\x(T)$ and the inverse map $f^{-1}$ as that from $\x(T)$ to $\x(0)$. In the
implementation, the dynamic system~\eqref{eq:ode} is realized by the fourth
order Runge-Kutta scheme with a sufficiently small stepsize $\tau$.  During the
evaluation of $q(\y,T)$ for an arbitrary $\y$, we first compute the inverse map
$f^{-1}(\y)$ by solving~\eqref{eq:xode} from $t=T$ to 0 with $\x(T)=\y$, and
then solve~\eqref{eq:pode} from $t=0$ to $T$ with $q(\x(0),0)=p_0(f^{-1}(\y))$.
During sampling, we first draw a sample $\y$ from the initial distribution $p_0$
and then output $f(\y)$ by solving~\eqref{eq:xode} from $t=0$ to $T$ with
$\x(0)=\y$.

For a predetermined time horizon $T$, the flow guided by different potential
functions $\phi(\x)$ may evolve the initial density $q(\x,0)=q_0(\x)$ into a
variety of densities $q(\x,T)$ at time $T$. From the perspective of optimal
control theory, the optimal potential function $\phi(\x)$ for approximating the
target density $q_1(\x)$ should be the solution to the following optimization
problem:
\begin{equation}
	\min_{\phi:\R^d\rightarrow\R} \mathrm{D}\left(q_1(\cdot),q(\cdot,T)\right),
	\label{eq:optcontrol}
\end{equation}
where $\mathrm{D}(\cdot,\cdot)$ is a proper metric or divergence for probability
measures. 

\begin{figure}[!htbp]
	\centering
	\begin{tikzpicture}[mybox/.style={draw, inner sep=5pt}]
		\node[mybox,label={left:{\bf\small\begin{tabular}{c} Density\\
		(Eulerian)\end{tabular}}}](box1) at (0,-1.5){%
			\begin{tikzcd}
				q(\cdot,0) = q_0(\cdot)  \arrow[r,"\eqref{eq:continuity}"]&
				q_\theta(\cdot)\equiv q(\cdot,T) \approx q_1(\cdot) 
			\end{tikzcd}
		}; \node[mybox,label={left:{\bf\small\begin{tabular}{c} Trajectories\\
		(Lagrangian)\end{tabular}}}](box2) at (0,1) {%
			\begin{tikzcd}
				q(\x(0),0) = q_0(\x(0))\arrow[r, "\eqref{eq:pode}", blue]&
				q_\theta(\x(T)) \equiv q(\x(T),T) \approx q_1(\x(T))\\ 
				\x(0) \arrow[r,shift left, "f~\eqref{eq:xode}", red]
				\arrow[u,"q_0", blue]  & \x(T) \arrow[l, shift left,
				"f^{-1}~\eqref{eq:xode}", blue]
			\end{tikzcd}
		}; \draw[<->, double] (box1)--(box2); \draw[->,red] (-1.7,-1.3) to[out =
		90, in = -90,red] node[midway,left]{\footnotesize sampling} (-3.5,0.05)
		;
	\end{tikzpicture}
	\caption{The mechanism of the continuous-time flow model: the blue path
		shows the evaluation procedure and the red arrow shows the sampling
		procedure for the resulting density $q_\theta(\cdot)$. }
	\label{fig:maflow}
\end{figure}
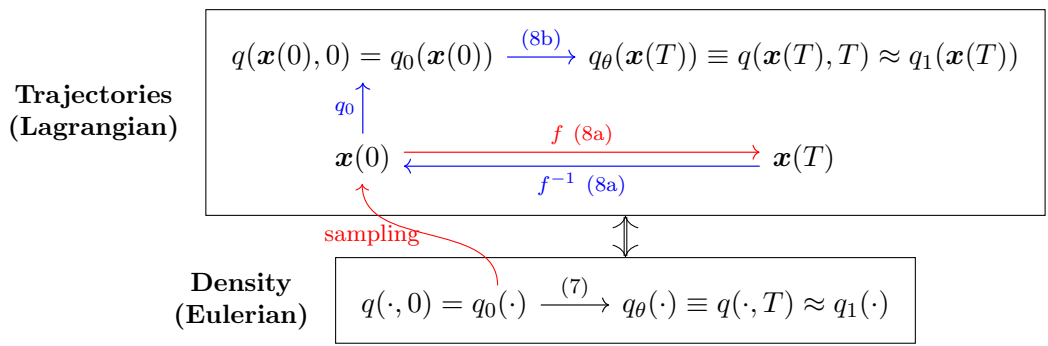

The potential function $\phi(\x)$ is parameterized by a neural network denoted
as $\phi_\theta(\x)$ with parameter $\theta$.  In what follows, we shall denote
the resulting pushforward $f$ and density $q(\x,T)$ by this continuous-time flow
model as $f_\theta$ and  $q_\theta(\x)$. Taking the metric
$\mathrm{D}(\cdot,\cdot)$ in~\eqref{eq:optcontrol} as the Kullback-Leibler (KL)
divergence, \eqref{eq:optcontrol} amounts to an MLE as in~\eqref{eq:MLE} and the
parameter $\theta$ of the neural network is thus trained by minimizing over the
negative log-likelihood
\[
	\theta
	=\argmin_\theta \E_{\x\sim q_1}\left[\log \dfrac{q_1(\x)}{q_\theta(\x)}\right]
	=\argmin_\theta \E_{\x\sim q_1}\left[-\log q_\theta(\x)\right].
\]
The mechanism of this continuous-time flow model is shown in
Figure~\ref{fig:maflow}.

\section{Tensorizing flow}
\label{sec:alg}

This section presents our tensorizing flow algorithm for high-dimensional
density estimation.  As an overview, the algorithm consists of two main steps:
\[
	\pE(\cdot)=\dfrac{1}{N}\sum_{i=1}^N \delta\left(\cdot-\x^{(i)}\right)
	\xrightarrow{1} p^\TT(\cdot) \xrightarrow{2} p^\TF(\cdot)
\]
\begin{enumerate}
	\item Construct an approximate tensor-train representation $p^\TT$ from the
	      samples $\{ \x^{(i)}\}_{1\le i\le N}$ by combining sketching
	      techniques with kernel density estimation (Section~\ref{sec:algstpA});
	\item Apply the continuous-time flow model to drive $p^\TT$ towards $\{
	      \x^{(i)}\}_{1\le i\le N}$, with the resulting distribution denoted by
	      $p^\TF$ (Section~\ref{sec:algstpB}).
\end{enumerate}

\subsection{Construction of $p^\TT$}
\label{sec:algstpA}

We start with a conceptual algorithm and follow by a more practical one.

\subsubsection{Ideal case}
\label{sec:ideal}

Let us motivate the construction of an approximate TT representation by
considering an ideal case where the underlying density $p$ has a {\em
finite-rank} structure and is also {\em Markovian}, which are explained as
follows.

We assume that all reshaped versions  $p(x_{1:k};x_{k+1:d})$ of $p(\x)$ for
$1\le k\le d-1$ are Hilbert-Schmidt kernels~\citep{stein2009real} so that we can
apply singular value decomposition (SVD)~\citep{young1988introduction} (also
called Schmidt decomposition) to them. For a Hilbert-Schmidt kernel $K$, we
define its \emph{column space} by its range, and its \emph{row space} by the
range of its adjoint.
\begin{definition}[Finite-rank]
	A probability density function $p(\x)$ is \emph{finite-rank} if for any
	$1\le k\le d-1$, the reshaped version $p(x_{1:k};x_{k+1:d})$ of $p(\x)$ as a
	Hilbert-Schmidt kernel is finite-rank, \emph{i.e.} of finite-dimensional
	column space.
	\label{def:finiterank}
\end{definition}

We also assume throughout that all the marginal distributions of $p(\x)$,
especially the 2 or 3-marginals $p_k$~\eqref{eq:marginals}, belong to the class
of Hilbert-Schmidt kernels so we can also perform SVD to them when necessary.

\begin{definition}[Markovian]
	A probability density function $p(\x)$ is \emph{Markovian} if it can be
	written in
	\begin{equation*}
		p(\x) = p(x_1)p(x_2|x_1)\cdots p(x_d|x_{d-1}).
	\end{equation*}
	\label{def:markov}
\end{definition}

\paragraph*{Finite-rank structure}

Under the finite-rank assumption, the cores of the TT representation of $p(\x)$
can be obtained simply via the following proposition:
\begin{proposition}[Core determining equation]
	\label{prop:cde}
	Suppose that the probability density $p$ is finite-rank. For $ 1\le k \le
	d-1$, denote the rank of its reshaped version $p(x_{1:k};x_{k+1:d})$ by
	$r_k$ and let $\{\Phi_k(x_{1:k};\alpha_k)\}_{1\le \alpha_k\le r_{k}}$ be the
	first $r_k$ left singular vectors of $p(x_{1:k};x_{k+1:d})$. Then there
	exists a unique solution $G_1:I\times[r_1]\rightarrow\R, G_2:[r_1]\times
	I\times[r_2]\rightarrow\R, \ldots, G_d:[r_{d-1}]\times I\rightarrow\R$ to
	the following system of {\em core determining equations (CDEs)}:
	\begin{equation}
		\begin{aligned}
			G_1(x_1;\alpha_1)                                                                               & =\Phi_1(x_1;\alpha_1),                           \\
			\sum_{\alpha_{k-1}=1}^{r_{k-1}}\Phi_{k-1}(x_{1:k-1};\alpha_{k-1})G_k(\alpha_{k-1};x_k,\alpha_k) & =\Phi_k(x_{1:k-1};x_k,\alpha_k),\ 2\le k\le d-1, \\
			\sum_{\alpha_{d-1}=1}^{r_{d-1}}\Phi_{d-1}(x_{1:d-1};\alpha_{d-1})G_d(\alpha_{d-1};x_d)          & =p(x_{1:d-1};x_d),                               
		\end{aligned}
		\label{eq:cde}  
	\end{equation}
	where the cores $G_k$ give an exact TT representation of $p(\x)$:
	\begin{equation}
		p(\x) = G_1(x_1,:) G_2(:,x_2,:)\cdots G_d(:,x_d).
		\label{eq:lowranktt}
	\end{equation}
\end{proposition}
We refer to Appendix~\ref{app:proof} for the proof of this proposition.

\paragraph*{Left-sketching technique}

Unfortunately, the size of equations in~\eqref{eq:cde} grows exponentially with
the dimension $d$ (even after discretization) and consequently it is impossible
to estimate all the coefficients $\Phi_k$ from finite samples. A key observation
is that this linear system is significantly over-determined so it can be reduced
efficiently by a sketching technique. To implement the sketching, we select
suitable left-sketching functions $S_{k-1}(y_{k-1};x_{1:k-1})$ for $2\le k\le
d$, where $y_{k-1}\in \Y_{k-1}$ with $\Y_{k-1}$ being an appropriate set
specified by the model.  By contracting them with the left-hand sides
of~\eqref{eq:cde}, we obtain the following reduced system of CDEs:
\begin{equation}
	\begin{aligned}
		G_1(x_1;\alpha_1)                                                                          & =B_1(x_1;\alpha_1),                         \\
		\sum_{\alpha_{k-1}=1}^{r_{k-1}}A_{k-1}(y_{k-1};\alpha_{k-1})G_k(\alpha_{k-1};x_k,\alpha_k) & =B_k(y_{k-1};x_k,\alpha_k),\ 2\le k\le d-1, \\
		\sum_{\alpha_{d-1}=1}^{r_{d-1}}A_{d-1}(y_{d-1};\alpha_{d-1})G_d(\alpha_{d-1};x_d)          & =B_d(y_{d-1};x_d),                          
	\end{aligned}
	\label{eq:sketchedcde}
\end{equation}
where the coefficients $B_k$ and $A_k$ are given by
\begin{equation}
	\begin{aligned}
		B_1(x_1; \alpha_1)=            & \Phi_1(x_1;\alpha_1),                                                                                  \\
		B_k(y_{k-1};x_k,\alpha_k)=     & \int_{I^{k-1}}S_{k-1}(y_{k-1};x_{1:k-1})\Phi_k(x_{1:k-1};x_k,\alpha_k)\d x_{1:k-1},\; 2\le k\le d-1    \\
		B_d(y_{d-1};x_d)=              & \int_{I^{d-1}}S_{d-1}(y_{d-1};x_{1:d-1})p(x_{1:d-1};x_d)\d x_{1:d-1},                                  \\
		A_{k-1}(y_{k-1};\alpha_{k-1})= & \int_{I^{k-1}}S_{k-1}(y_{k-1};x_{1:k-1})\Phi_{k-1}(x_{1:k-1};\alpha_{k-1})\d x_{1:k-1},\; 2\le k\le d. 
	\end{aligned}
	\label{eq:ab}  
\end{equation}

\begin{figure}[!htbp]
	\centering
	\begin{tikzpicture}
		\node[draw, fill = red, fill opacity=0.1, rectangle,
		fit={(-4.3,-0.8)(2.3,0.9)}, label={below:{$\Phi_{k-1}$}}]  {};
		\node[draw, fill = yellow, fill opacity=0.1, rectangle,
		fit={(-4.4,-1.4)(4.3,1.0)}, label={-20:{$\Phi_{k}$}}] (phi) {};
		\node[draw, fill = blue, fill opacity=0.1, rectangle,
		fit={(-4.5,-1.5)(2.4,2.9)}, label={above:{$A_{k-1}$}}]  {};
		        
		\node[circle,draw, scale=2,label= {below: $B_k$}](bk) at (6.6,0.3) {};
		\node[left = 0.4 of bk] (yk-1) {$y_{k-1}$}; \draw[dashed] (bk) --
		(yk-1); \node[above = 0.4 of bk] (xk)  {$x_{k}$}; \draw[dashed] (bk) --
		(xk); \node[right = 0.4 of bk] (alphak) {$\alpha_{k-1}$}; \draw (bk) --
		(alphak); \path (yk-1) -- node[auto=false]{$=$} (phi);
		            
		\node[circle,draw, scale=2,label= {below: $G_1$}](1) at (-4,0) {};
		\draw[dashed] (1) --node[label={[xshift = -3pt, yshift =
		-3pt]right:$x_1$}]{} (-4,1.38); \node[circle,draw, scale=2,label=
		{below: $G_2$}](2) at (-2,0) {};
		\draw[dashed] (2) --node[label={[xshift = -3pt, yshift =
		-3pt]right:$x_2$}]{} (-2,1.38); \node[circle,draw, scale=2,label=
		{below: $G_{k-1}$}](k-1) at (1,0) {};
		\draw[dashed] (k-1) --node[label={[xshift = 3pt, yshift =
		-3pt]left:$x_{k-1}$}]{} (1,1.38); \path (2)-- node[auto=false]{$\cdots$}
		(k-1); \node[circle,draw, scale=2,label= {below: $G_{k}$}](k) at (3.1,0)
		{}; \node (ka) at (3.1,0.8) {$x_k$}; \draw[dashed] (k) -- (ka);
		            
		\node[draw, rounded rectangle,
		fit={(-4.2,1.5)(1.2,1.9)},label={above:$S_{k-1}$}](B) {}; \node (y) at
		(1,2.7) {$y_{k-1}$}; \draw[dashed] (1,2.01) -- (y);
		            
		\draw (1)--node[label={[yshift = -3pt]above:$\alpha_1$}]{}(2); \draw
		(2)--(-1,0); \draw (0,0)--(k-1); \draw (k-1)--node[label={[xshift =
		-5pt, yshift = -3pt]above:$\alpha_{k-1}$}]{}(k); \node[right of = k]
		(alpha)  {$\alpha_k$}; \draw (k)--(alpha);
	\end{tikzpicture}
	\caption{The diagrammatic notation of the $k$-th equation in the reduced
	CDEs~\eqref{eq:sketchedcde} for $2\le k\le d-1$ (\emph{cf.} the corresponding equation in the original CDEs~\eqref{eq:cde}). }
	\label{fig:cde}
\end{figure}
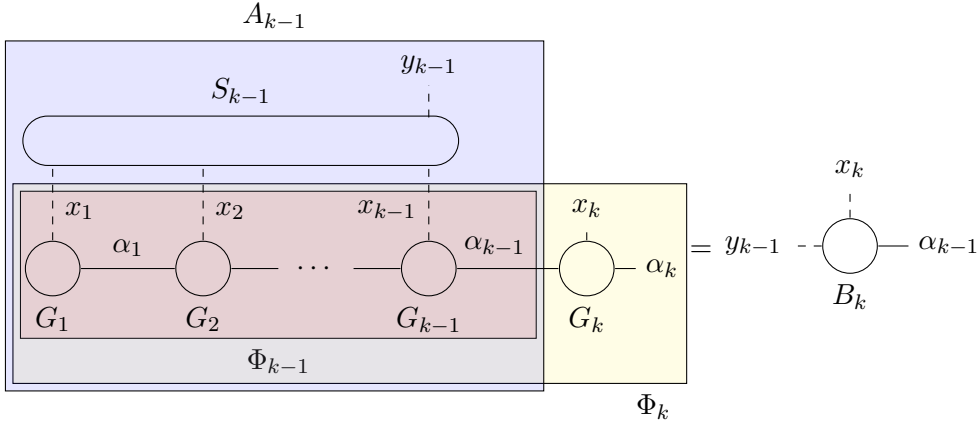

Generally, the left-sketching functions $S_{k-1}(y_{k-1};x_{1:k-1})$ need to be
chosen such that the row space of $\Phi_k(x_{1:k-1};x_k,\alpha_k)$ is retained
and the variance of the coefficient matrices is reduced as much as
possible~\citep{hur2022generative}. For an illustration of the sketching
technique, we refer readers to Figure~\ref{fig:cde} for the diagrammic notation
of the $k$-th equation in the reduced CDEs~\eqref{eq:sketchedcde} for $2\le k
\le d-1$ (\emph{cf.} the corresponding equation in the original
CDEs~\eqref{eq:cde}).

\paragraph*{Markovian structure}

In general, it is unclear what $S_k$ one needs to choose in order to obtain
$B_k$ and $A_k$ in~\eqref{eq:ab}. Furthermore, in practice it is impossible to
compute or estimate the singular vectors $\Phi_k$ involved. However, under the
extra Markovian assumption, the computation for $B_k$ and $A_k$ can be made
simple due to the following lemma:
\begin{lemma}[{\citet[Lemma 5]{hur2022generative}}] Suppose $p(\x)$ is
	Markovian, then for any $i\le k \le j-1$,
	\begin{enumerate}
		\item $p(x_{i:k};x_{k+1:j})$ and $p(x_{i:k};x_{k+1})$ have the same
		column space;
		\item $p(x_{i:k};x_{k+1:j})$ and $p(x_k;x_{k+1:j})$ have the same row
		space.
	\end{enumerate}
	\label{lem:markov}
\end{lemma}

Lemma~\ref{lem:markov} essentially tells us that for $p(x_{1:k};x_{k+1:d})$,
marginalizing out $x_{k+2:d}$ or $x_{1:k-1}$ will not affect the corresponding
column or row space.  Motivated by this lemma, we are able to make the following
two simplifications when computing the coefficients $B_k$ and $A_k$
in~\eqref{eq:cde}: 
\begin{enumerate}
	\item Obtain $\Phi_k(x_{1:k};\alpha_{k})$ for $1\le k\le d-1$ by only
	      considering the column space of the $(k+1)$-dimensional marginal
	      distribution $p(x_{1:k};x_{k+1})$ instead of the full distribution
	      $p(x_{1:k};x_{k+1:d})$;
	\item Simply take $\Y_{k-1} = I$ and $S_{k-1}(y_{k-1};x_{1:k-1}) =
	      \delta(y_{k-1}-x_{k-1})$, \emph{i.e.} the Schwartz distribution that
	      marginalizes out the first $k-2$ dimensions, for $2\le k \le d$ as
	      suggested by~\citet{hur2022generative}.
\end{enumerate}

For $k=1$, these simplifications indicate that $\Phi_1(x_1;\alpha_1)$ can be
obtained directly by applying SVD to the 2-marginal $p_1(x_1;x_2)$, and
subsequently
\[
	A_1(y_1;\alpha_1)=\int_{I}\delta(y_{1}-x_{1})\Phi_{1}(x_{1};\alpha_{1})\d
	x_{1}=\Phi_1(y_1;\alpha_1)=B_1(y_1;\alpha_1),
\]
where the last equality is by definition~\eqref{eq:ab}. Similarly, for $k=d$,
$B_d(y_{d-1};x_d)=p_d(y_{d-1};x_d)$. 

For $2\le k \le d-1$, the simplifications yield
\begin{equation}
	\begin{aligned}
		B_k(y_{k-1};x_k,\alpha_k)
		=&\int_{I^{k-1}}\delta(y_{k-1}-x_{k-1})\Phi_k(x_{1:k-1};x_k,\alpha_k)\d
         x_{1:k-1}
		  \\ =&\int_{I^{k-2}}\Phi_k(x_{1:k-2};y_{k-1},x_k,\alpha_k)\d x_{1:k-2}.
	\end{aligned}
	\label{eq:bk}
\end{equation}
A natural way to obtain $B_k$ is to first calculate $\Phi_k(x_{1:k};\alpha_{k})$
by performing SVD directly to $p(x_{1:k};x_{k+1})$ and apply left-sketching
afterwards, \emph{i.e.} marginalizing out $x_{1:k-2}$ from
$\Phi_k(x_{1:k};\alpha_{k})$ as in~\eqref{eq:bk}. However, this approach is
practically infeasible, since $p(x_{1:k};x_{k+1})$ is again exponentially large
to $d$ and its range $\Phi_k(x_{1:k};\alpha_{k})$ can hardly be estimated by a
limited collection of samples. Thus instead, we obtain $B_k$ in an implicit
manner by first applying the left-sketching functions $S_{k-1}$ to
$p(x_{1:k};x_{k+1})$, \emph{i.e.} marginalizing out $x_{1:k-2}$ from
$p(x_{1:k};x_{k+1})$ to obtain the 3-marginal $p_k(x_{k-1},x_k;x_{k+1})$, and
then performing SVD to $p_k(x_{k-1},x_k;x_{k+1})$. Then
$B_k(x_{k-1},x_k;\alpha_k)$ is formed by the first $r_k$ left singular vectors
of $p_k(x_{k-1},x_k;x_{k+1})$.  
Moreover, since
\begin{equation*}
	\begin{aligned}
		  & A_{k}(y_{k};\alpha_{k})=\int_{I^{k}}\delta(y_{k}-x_{k})\Phi_{k}(x_{1:k};\alpha_{k})\d x_{1:k}=\int_{I^{k-1}}\Phi_{k}(x_{1:k-1};y_k,\alpha_{k})\d x_{1:k-1} \\
		  & =\int_I \int_{I^{k-2}}\Phi_{k}(x_{1:k-2};y_{k-1},y_k,\alpha_{k})\d x_{1:k-2}\d y_{k-1}      =\int_{I}B_{k}(y_{k-1};y_k,\alpha_{k})\d y_{k-1},              
	\end{aligned}
\end{equation*}
$A_k$ is obtained subsequently by marginalizing out the first dimension of
$B_k$.

In conclusion, we are able to obtain an exact TT representation in the form
of~\eqref{eq:lowranktt} for the finite-rank and Markovian distributions $p(\x)$
by first forming the coefficients $B_k$ and $A_k$ and then solving the reduced
system of CDEs~\eqref{eq:sketchedcde}.

\subsubsection{General case}
\label{sec:general}

The ideal case in Section~\ref{sec:ideal} assumes that the distribution $p$ is
finite-rank and Markovian. It also assumes the function access to the marginals
$p_k$~\eqref{eq:marginals} and the singular value decomposition for
Hilbert-Schmidt kernels (rather than finite-dimensional matrices). In practice,
we need to deal with an unknown underlying distribution $p^*$, which is not
necessarily low-rank and Markovian. Furthermore, instead of the analytic
formulae of the marginals, we only have access to a limited set of samples
$\{\x^{(i)}\}_{1 \le i\le N}$ drawn from the unknown $p^*$.

In order to bridge this gap between the ideal case and the practical situation,
we describe below how to adapt the method described in Section~\ref{sec:ideal}.
The resulting TT $p^{\TT}$ will serve as a reasonable approximation to the
unknown underlying distribution $p^*$.

\paragraph*{Step 1.}

Construct the kernel density estimators $\pS_k$ of the marginals
$p^*_k$~\eqref{eq:marginals} from the samples $\{ \x^{(i)} \}_{1\le i\le N}$ for
$1\le k\le d$.

When evaluating the coefficients $B_k$ and $A_k$, the most direct approach is to
estimate the marginals $p_k^*$ by directly interpolating the marginal
distribution $\pE_k$ of the empirical distribution $\pE$~\eqref{eq:emp} with
polynomials as in the work by~\citet{hur2022generative}. Instead, we estimate
$p_k^*$ by applying kernel density estimation (KDE) to the corresponding slices
of samples, \emph{e.g.} for $2\le k \le d-1$, $p_k^*$ is estimated by the kernel
density estimators 
\[
	\pS_k(x_{k-1:k+1}):=\dfrac{1}{Nh}\sum_{i=1}^N K\left( \dfrac{x_{k-1:k+1}-x^{(i)}_{k-1:k+1}}{h} \right),
\]
where $K(\cdot)$ is the Gaussian kernel $(2\pi)^{-3/2}\exp\left(-\|\cdot\|^2/2
\right)$ and $h$ is the bandwidth.

\begin{remark}
	The adoption of KDE to marginal distributions is key to the construction of
	the TT representation in practice. The reason is that if we perform SVD
	directly to the marginal empirical distributions $\pE_k$, the sparsity of
	samples may lead to severe Gibbs phenomenon in the resulting approximate TT
	representation, posing a major obstacle in the way of implementing
	tensorizing flow.  Therefore, $\pE_k$ are first smoothed by KDE and SVD is
	performed to the kernel density estimators $\pS_k$ instead of $\pE_k$.  It
	is also noteworthy that KDE is performed only for estimating the 2 or
	3-marginals $p^*_k$ but \textit{not} the full distribution $p^*$, for
	applying KDE directly to $p^*$ would lead to poor performance because of the
	curse of dimensionality.
	
	In general, there is a bias-variance trade-off for choosing the bandwidth
	parameter $h$. Specifically, when $h\rightarrow 0$, the kernel density
	estimator $\hat{p}_k$ approaches the empirical distribution $\pE_k$, an
	unbiased estimator of the true distribution $p_k$. As the bandwidth $h$
	grows, $\hat{p}_k$ becomes smoother with certain bias. When the bandwidth
	$h$ is sufficiently large, $\hat{p}_k$ is smooth enough to be well
	approximated by polynomial approximation.
\end{remark}

\paragraph*{Step 2.}

Estimate the coefficients $B_k$ for $1\le k\le d$ and $A_k$ for $1\le k\le d-1$
from the kernel density estimators $\pS_k$.

Ideally, $B_d = \pS_{d}$ and for $1\le k\le d-1$, $B_k$ is formed by the first
$r_k$ left singular vectors by performing SVD to the $d-1$ kernel density
estimators
\[
	\pS_1(x_1;x_2),\; \pS_2(x_{1},x_{2};x_{3}),\ldots,\pS_{d-1}(x_{d-2},x_{d-1};x_{d}).
\]
Afterwards, $A_1=B_1$, and $A_k$ is obtained by marginalizing out the first
variable of $B_k$ for $2\le k\le d-1$. 

However, since $x_i$ takes value in $I=[-1,1]$ and all marginals are continuous
functions, numerical approximation is needed for carrying out SVD practically.
To this end, we introduce the normalized Legendre polynomials
$\{L_{i}(x)\}_{i\ge 1}$ with $\text{deg}(L_i) = i-1$, which form an orthonormal
basis of $L^2(I)$. For example, when evaluating $B_k$ for $2\le k\le d-1$, we
take the tensor-product normalized Legendre polynomials
$\{L_{i_{k-1}}(x_{k-1})L_{i_k}(x_k)\}_{1\le i_{k-1},i_k \le M}$ as the expansion
basis for variables $(x_{k-1},x_k)$ and $\{L_{i_{k+1}}(x_{k+1})\}_{1\le
i_{k+1}\le M}$ for $x_{k+1}$. Here $M$ is a constant that controls the accuracy
of the polynomial approximation. Projecting $\pS_k(x_{k-1},x_k;x_{k+1})$
orthogonally onto these two sets of basis functions gives the following $M^2
\times M$ coefficient matrix with entry
\[
	\begin{aligned}
		  & \mathsf{P}^S_k(i_{k-1},i_k; i_{k+1}) \\=& \int_{I\times I} \int_I \left(L_{i_{k-1}}(x_{k-1}) L_{i_k}(x_k)\right) \pS_k(x_{k-1},x_k;x_{k+1})   L_{i_{k+1}}(x_{k+1}) \d x_{k-1} \d x_k \d x_{k+1}.
	\end{aligned}
\]
Next, one computes the truncated SVD for $\mathsf{P}^S_k(i_{k-1},i_k; i_{k+1})$
and groups the first $r_k$ singular vectors into a matrix
$\B_k(i_{k-1},i_k;\alpha_{k})$ of size $M^2 \times r_k$, where $r_k$ is the
numerical rank. Finally, $B_k(x_{k-1},x_k;\alpha_{k})$ can be obtained by
\begin{equation*}
	B_k(x_{k-1},x_{k};\alpha_{k})
	:= \sum_{i_{k-1}=1}^M \sum_{i_{k}=1}^M  \B_k(i_{k-1},i_k;\alpha_k) L_{i_{k-1}}(x_{k-1}) L_{i_k}(x_k),
\end{equation*}
and by contracting $L_1(x_{k-1})\equiv 1/\sqrt{2}$ to both sides, $A_k$ is
obtained subsequently by
\begin{equation*}
	A_k(x_{k};\alpha_{k})
	:= \sqrt{2}\sum_{i_{k}=1}^M  \B_k(1,i_k;\alpha_k) L_{i_k}(x_k).
\end{equation*}
The cases for $k=1$ and $d$ are handled similarly.

\paragraph*{Step 3.}

Solve~\eqref{eq:sketchedcde} by least squares for the cores $G_1,\ldots,G_d$.

Similar to the previous step, since \eqref{eq:sketchedcde} is formulated in
terms of functions, numerical approximation is needed. We again resort to
polynomial approximation and expand $G_k$, $B_k$, and $A_k$ w.r.t. the first $M$
normalized Legendre polynomials, \emph{e.g.} for $2\le k\le d-1$, the
corresponding coefficient matrices $\G_k$, $\B_k$, and $\A_k$ are given by
\begin{equation*}
	\begin{aligned}
		\G_k(\alpha_{k-1};i_k,\alpha_k)    & =\int_{I} G_k(\alpha_{k-1};x_k,\alpha_k)L_{i_k}(x_k)\d x_k,                                            \\
		\B_k(\beta_{k-1};i_k,\alpha_k)     & =\int_{I\times I} B_k(y_{k-1}; x_k, \alpha_k)  L_{\beta_{k-1}}(y_{k-1}) L_{i_k}(x_k) \d y_{k-1} \d x_k \\
		\A_{k-1}(\beta_{k-1};\alpha_{k-1}) & =\int_I A_{k-1}(y_{k-1};\alpha_{k-1})  L_{\beta_{k-1}}(y_{k-1}) \d y_{k-1}.                            
	\end{aligned}
	\label{eq:tildeG}
\end{equation*}

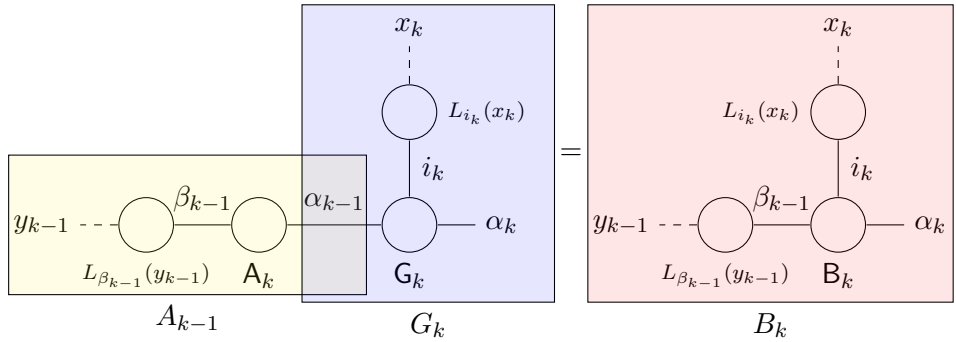
\begin{figure}[!htbp]
	\centering
	\begin{tikzpicture}
		\node[circle,draw, scale=2,label= {below: $\B_k$}](bk) at (7.2,0) {};
		\node[circle,draw, scale=2,label= {left: {\scriptsize
		$L_{i_{k}}(x_{k})$}}](bu) at (7.2,1.5) {}; \node[circle,draw,
		scale=2,label= {below: {\scriptsize $L_{\beta_{k-1}}(y_{k-1})$}}](bl) at
		(5.7,0) {}; \node[left = 0.5 of bl] (bll) {$y_{k-1}$}; \node[above = 0.5
		of bu] (buu) {$x_{k}$}; \node[right = 0.5 of bk] (br) {$\alpha_{k}$};
		\draw[dashed] (bu) -- (buu); \draw[dashed] (bl) -- (bll); \draw (bk) --
		(br); \draw (bk) --node[label={[yshift = -3pt]above:{$\beta_{k-1}$}}]{}
		(bl); \draw (bk) --node[label={[xshift = -3pt]right:{$i_{k}$}}]{} (bu);
		        
		\node[draw, fill = red, fill opacity=0.1, rectangle,
		fit={(4.0,-0.9)(8.6,2.8)}, label={below:{$B_{k}$}}](Bk){};
		        
		\node[circle,draw, scale=2,label= {below: $\G_k$}](gk) at (1.5,0) {};
		\node[circle,draw, scale=2,label= {right: {\scriptsize
		$L_{i_{k}}(x_{k})$}}](gu) at (1.5,1.5) {}; \node[circle,draw,
		scale=2,label= {below: {$\A_k$}}](ak) at (-0.5,0) {}; \node[circle,draw,
		scale=2,label= {below: {\scriptsize $L_{\beta_{k-1}}(y_{k-1})$}}](al) at
		(-2,0) {}; \node[left = 0.5 of al] (all) {$y_{k-1}$}; \node[above = 0.5
		of gu] (guu) {$x_{k}$}; \node[right = 0.5 of gk] (gr) {$\alpha_{k}$};
		\draw[dashed] (gu) -- (guu); \draw[dashed] (al) -- (all); \draw (gk) --
		(gr); \draw (ak) --node[label={[yshift = -3pt]above:{$\beta_{k-1}$}}]{}
		(al); \draw (gk) --node[label={[yshift = -3pt]above:{$\alpha_{k-1}$}}]{}
		(ak); \draw (gk) --node[label={[xshift = -3pt]right:{$i_{k}$}}]{} (gu);
		        
		\node[draw, fill = yellow, fill opacity=0.1, rectangle,
		fit={(-3.7,-0.8)(0.8,0.8)}, label={below:{$A_{k-1}$}}](Ak){};
		\node[draw, fill = blue, fill opacity=0.1, rectangle,
		fit={(0.2,-0.9)(3.3,2.8)}, label={below:{$G_{k}$}}](Gk){};
		        
		\path (Bk) -- node[auto=false]{=} (Gk);
		        
	\end{tikzpicture}
	\caption{The diagrammatic notation of the $k$-th equation in the discrete
	CDEs~\eqref{eq:discretecde} for $2\le k\le d-1$ (\emph{cf.} the corresponding equation in the reduced CDEs~\eqref{eq:sketchedcde}). }
	\label{fig:discretecde}
\end{figure}

As interpreted by diagrammic notation in Figure~\ref{fig:discretecde}, the
projected version of the system~\eqref{eq:sketchedcde} is
\begin{equation}
	\begin{aligned}
		\G_1(i_1;\alpha_1)&=\B_1(i_1;\alpha_1),\\ \sum_{\alpha_{k-1}=1}^{r_{k-1}}
		\A_{k-1}(\beta_{k-1};\alpha_{k-1})\G_k(\alpha_{k-1};i_k,\alpha_k) & =\B_k(\beta_{k-1};i_k,\alpha_k),\ 2\le k \le d-1, \\ \sum_{\alpha_{d-1}=1}^{r_{d-1}}\A_{d-1}(\beta_{d-1};\alpha_{d-1})\G_d(\alpha_{d-1};i_d)&=\B_d(\beta_{d-1};i_d).
	\end{aligned}
	\label{eq:discretecde}  
\end{equation}
The discrete cores $\G_k$ can be solved efficiently from these equations by
applying least squares. Once $\G_k$ are solved, they are combined with the
normalized Legendre polynomials to produce the continuous cores
\[
	G_k(\alpha_{k-1};x_k,\alpha_k) \approx \sum_{i_k=1}^M \G_k(\alpha_{k-1};i_k,\alpha_k) L_{i_k}(x_k).
\]

\paragraph*{Step 4.}

With the cores $G_k$ ready, the approximate TT representation $p^\TT(\x)$ of
$\pE(\x)$ can be set to ${G}_1(x_1,:) {G}_2(:,x_2,:)\cdots {G}_d(:,x_d)$ as
in~\eqref{eq:lowranktt}. However, there are two extra issues to be addressed.

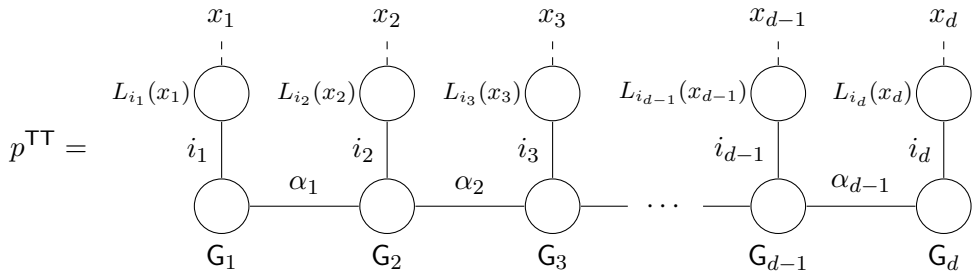
\begin{figure}[!htbp]
	\centering
	\begin{tikzpicture}
		\node[circle,draw, scale=2,label= {below: $\G_1$}](1) at (-4.4,0) {};
		\node at (-6.7,0.8) {$p^\TT=$}; \node[circle,draw, scale=2,label=
		{below: $\G_2$}](2) at (-2.2,0) {}; \node[circle,draw, scale=2,label=
		{below: $\G_3$}](3) at (0,0) {}; \node[circle,draw, scale=2,label=
		{below: $\G_{d-1}$}](d-1) at (3,0) {}; \node[circle,draw, scale=2,label=
		{below: $\G_d$}](d) at (5.2,0) {};
		        
		\node[circle,draw, scale=2,label= {[xshift=3pt]-180:{\footnotesize
		$L_{i_1}(x_1)$}}] (11) at (-4.4,1.5) {}; \node[circle,draw,
		scale=2,label= {[xshift=3pt]-180:{\footnotesize $L_{i_2}(x_2)$}}] (22)
		at (-2.2,1.5) {}; \node[circle,draw, scale=2,label=
		{[xshift=3pt]-180:{\footnotesize $L_{i_3}(x_3)$}}] (33) at (0,1.5) {};
		\node[circle,draw, scale=2,label= {[xshift=3pt]-180:{\footnotesize
		$L_{i_{d-1}}(x_{d-1})$}}] (dd-1) at (3,1.5) {}; \node[circle,draw,
		scale=2,label= {[xshift=3pt]-180:{\footnotesize $L_{i_d}(x_d)$}}] (dd)
		at (5.2,1.5) {};
		        
		\node[above of = 11]  (1aa) {$x_1$}; \node[above of = 22]  (2aa)
		{$x_2$}; \node[above of = 33]  (3aa) {$x_3$}; \node[above of = dd-1]
		(d-1aa) {$x_{d-1}$}; \node[above of = dd]  (daa) {$x_d$};
		        
		\draw (1)--node[label={[yshift = -3pt]above:$\alpha_1$}]{} (2); \draw
		(2)--node[label={[yshift = -3pt]above:$\alpha_2$}]{}(3); \draw
		(3)--(1,0); \path (3)-- node[auto=false]{$\cdots$} (d-1); \draw
		(2,0)--(d-1); \draw (d-1)--node[label={[yshift =
		-3pt]above:$\alpha_{d-1}$}]{}(d);
		        
		\draw (1) --node[label={[xshift = 3pt]left:$i_1$}]{} (11); \draw[dashed]
		(11) -- (1aa); \draw (2) --node[label={[xshift = 3pt]left:$i_2$}]{}
		(22); \draw[dashed] (22) -- (2aa); \draw (3) --node[label={[xshift =
		3pt]left:$i_3$}]{} (33); \draw[dashed] (33) -- (3aa); \draw
		(d-1)--node[label={[xshift = 3pt]left:$i_{d-1}$}]{} (dd-1);
		\draw[dashed] (dd-1) -- (d-1aa); \draw (d) --node[label={[xshift =
		3pt]left:$i_d$}]{} (dd); \draw[dashed] (dd) --  (daa);
	\end{tikzpicture}
	\caption{The diagrammatic notation of the approximate TT representation
	$p^\TT$. }
	\label{fig:expandedtt}
\end{figure}

First, ${G}_1(x_1,:) {G}_2(:,x_2,:)\cdots {G}_d(:,x_d)$ does not necessarily
integrate to unity. The normalization can be done by contracting
${G}_1(x_1,:){G}_2(:,x_2,:)\cdots{G}_d(:,x_d)$ with the all-one function and
absorbing the resulting constant into any of $G_k$ s.t. $p^\TT$ retains the form
\begin{equation}
	p^\TT(x_{1:d}) :=  {G}_1(x_1,:) {G}_2(:,x_2,:)\cdots {G}_d(:,x_d),
	\label{eq:tildeP}
\end{equation}
the diagrammic notation of which is shown in Figure~\ref{fig:expandedtt}.

The second issue is that ${G}_1(x_1,:) {G}_2(:,x_2,:)\cdots {G}_d(:,x_d)$ is not
necessarily non-negative. To ensure the non-negativity, we can adopt the
following post-processing by following the approach of \textit{Born
machine}~\citep{han2018unsupervised}, \emph{i.e.} one solves
\[
	\begin{gathered}
		\min_{q(\x)} \left\| p^\TT(\x) - r(\x)^2\right\|_{L^2(I^d)}\\
		\textrm{s.t.  }r(\x)=\sum_{i_1=1}^{M}\cdots\sum_{i_{d}=1}^{M}\mathsf{R}(i_{1:M})L_{i_1}(x_1)\cdots L_{i_d}(x_d),
	\end{gathered}
\]
where $\mathsf{R}(i_{1:d}) = \mathsf{H}_1(i_1,:) \mathsf{H}_2(:,i_2,:) \cdots
\mathsf{H}_d(:,i_d)$ is a discrete tensor-train with discrete cores
$\mathsf{H}_i$. Noticing that
\[
	\int_{I^d} p^\TT(\x) \d \x=\int_{I^d} r(\x)^2 \d \x=\|\mathsf{R}\|_F^2
\]
by the orthogonality of Legendre polynomials, then
\begin{equation}
	p^\TT(\x) := r(\x)^2
	\label{eq:pq2}
\end{equation}
is guaranteed to be non-negative and integrate to one by normalizing the
Frobenius norm of the discrete tensor-train $\mathsf{Q}$. Strictly speaking,
this is not a TT representation, rather the pointwise square of a TT
representation.

\subsection{Construction of $p^\TF$}
\label{sec:algstpB}

In the second step of our method, instead of using the normal distribution as
the base distribution in normalizing flow, we start from the approximate TT
representation $p^\TT(\x)$ obtained in~\eqref{eq:tildeP} or~\eqref{eq:pq2} and
use the continuous-time flow model in Section~\ref{sec:maflow} to improve this
approximation.

Following Section~\ref{sec:maflow}, we choose the initial distribution
$q(\x,0)=p^\TT(\x)$, and then select a proper time horizon $T$ and stepsize
$\tau$ to obtain a new density approximation $q_\theta(\x) \equiv q(\x, T)$,
where the subscript $\theta$ indicates the neural network used to parameterize
the potential function $\phi_\theta(\x)$ that guides the flow~\eqref{eq:ode}.

As in the MLE setup~\eqref{eq:MLE}, the loss function for training is chosen as
the negative log-likelihood:
\begin{equation}
	\mathcal{L}(\theta)  
	:= - \E_{\x\sim \pE}\log q_\theta(\x).
	\label{eq:loss}
\end{equation}
In the actual implementation, the neural network is trained on batches. Each
batch is randomly selected from the full sample set $\{ \x^{(i)} \}_{1\le i\le
N}$, and the loss function is approximated by
$-1/N_{\text{batch}}\sum_{j=1}^{N_{\text{batch}}}\log q_\theta(\x^{(j)})$ within
the batch in each step, where $N_{\text{batch}}$ is the batch size. Each
likelihood $q_\theta(\x^{(j)})$ is calculated by solving the dynamic
system~\eqref{eq:ode} by the fourth order Runge-Kutta scheme.

Once $q_\theta(\x)$ is learned, we define the final product 
\begin{equation*}
	p^\TF(\x) := q_\theta(\x)
\end{equation*}
that can serve as an approximation to the unknown underlying distribution
$p^*(\x)$. Sampling from $p^\TF(\x)=q_\theta(\x)$ is carried out by first
sampling from the approximate TT representation $p^\TT(\x)$\footnote{One may
refer to the algorithms by~\citet{dolgov2020approximation}
and~\citet{novikov2021tensor} for efficiently sampling from a given TT
representation.} and then applying the pushforward $f$ again by numerically
integrating~\eqref{eq:xode}. Readers may refer to Figure~\ref{fig:maflow} for
the evaluation and sampling procedures for $q_\theta$.

\begin{remark}
	In the case of normalizing flow, the base distribution chosen as a normal
	distribution has no information of the target distribution $p^*$. Thus, a
	large neural network is needed so that the flow model is sufficiently
	expressive to learn the complicated pushforward from the normal distribution
	to $p^*$. However, in our tensorizing flow approach, the base distribution
	$p^\TT$ chosen as the TT representation is already believed to approximate
	$p^*$ well so the flow model here suffices to be close to the identity map,
	and it is expected to learn a good density approximation $p^\TF$ with a
	simple and easy-to-train neural network, which may lead to better
	generalization for our model as well. Furthermore, when the potential
	function $\phi_\theta$ in the continuous-time flow model is initialized as a
	constant function, both the forward and inverse map are obviously the
	identity at the beginning of training. It means that $p^\TT$ as an initial
	approximation of $\pE$ is exploited as priori knowledge, and we are
	guaranteed to obtain a better density approximation $p^\TF$ than $p^\TT$
	through training.   

	For this near-identity flow, one may also consider adopting residual
	flows~\citep{he2016deep}. Several related works
	\citep[\emph{e.g.}][]{gomez2017reversible,jacobsen2018revnet} use some
	techniques by introducing extra variables to create reversible network
	architectures based on residual connections.  However, compared with several
	classical flows such as NICE~\citep{dinh2014nice}, Real
	NVP~\citep{dinh2016density}, MAF~\citep{papamakarios2017masked}, and
	Glow~\citep{kingma2018glow}, these networks cannot be inverted analytically,
	which greatly affects their efficiency and feasibility. Moreover, due to the
	use of convolutional layers in these networks, the evaluation of the
	Jacobian determinant in~\eqref{eq:jacob} is very expensive and often
	requires a biased yet still expensive estimate of the log-Jacobian given by
	the power series for the trace of the matrix logarithm
	$\log\left(\det(I+F)\right)=\mathrm{tr}\left(\log(I+F)\right)=\sum_{k=1}^\infty(-1)^{k-1}\mathrm{tr}(F)^k/{k}$.
	Since we are using an ODE-based continuous-time flow model, both the path
	and the log-Jacobian can be obtained by numerical integration
	of~\eqref{eq:ode}, circumventing the inefficiency aforementioned.
\end{remark}

Before ending the algorithmic discussion, we provide below a summary of our
method.
\begin{algorithm}[!htbp]
	\caption{Tensorizing flow}\label{alg}
	\begin{algorithmic}
		\REQUIRE A collection of samples $\{ \x^{(i)}  \}_{1\le i\le N}$
		independently drawn from an underlying distribution
		$p^*(\x):I^d\rightarrow\R$;
		    
		\begin{enumerate}[leftmargin=*]
			\item Construct the approximate TT representation $p^\TT(\x)$ from
			      the samples $\{ \x^{(i)} \}_{1\le i\le N}$ following the
			      routine outlined in Section~\ref{sec:general}.
			\item Construct a potential function $\phi_\theta(\x)$ parameterized
			      by the neural network $\theta$, set $q(\x,0)=p^\TT(\x)$, and
			      construct the density estimation $q_\theta(\x)=q(\x,T)$ by
			      applying Runge-Kutta scheme to~\eqref{eq:ode} with stepsize
			      $\tau$ for $\lfloor T/\tau \rfloor$ steps;
			\item Train the neural network on the sample set $\{ \x^{(i)}
			      \}_{1\le i\le N}$ w.r.t. loss function~\eqref{eq:loss} and
			      output $q_\theta(\x)$ as the final estimation $p^\TF(\x)$ for
			      $p^*(\x)$.
		\end{enumerate}
		
	\end{algorithmic}
\end{algorithm}

\section{Experimental results}
\label{sec:experiments}

We present here several experimental results that illustrate the performance of
our algorithm. Under the assumption $\text{supp}(p)\subset I^d$, the algorithm
is implemented with proper transformation and scaling of the Legendre
polynomials for an arbitrary interval $I$ other than $[-1,1]$, and we will
specify the choice of $I$ for each example below. Gauss-Legendre quadrature is
adopted for all numerical integration involved in the construction of the TT
representation with $l$ quadrature points along each dimension. 

For the neural network used to parameterize the potential function
$\phi_\theta:\R^d\rightarrow\R$ in the flow model, we adopt a multi-layer
perceptron (MLP) structure with an input layer, two hidden layers of $D$
neurons, and an output layer.  The activation functions are chosen as
$\log\cosh$ and the softplus function for the first and second hidden layer,
respectively, in order to provide sufficient smoothness for $\phi_\theta$ as
well as $q_\theta$.  We use the Adam optimizer for the training of the neural
network with two parameters: learning rate (LR) and weight decay (WD), and scale
the learning rate by a multiplicative factor $\gamma$ after each epoch, the
choices of which are organized in Table~\ref{tab:hyper} in
Appendix~\ref{app:hyper}. All the experiments are implemented using PyTorch deep
learning framework and conducted by a Tesla V100 GPU.

Throughout all the examples in this section, we present the experiment results
of our tensorizing flow (TF) algorithm (Algorithm~\ref{alg}) compared with those
of normalizing flow (NF), where we use the continuous-time flow model of the
same neural network architecture and parameters (see Table~\ref{tab:hyper}) as a
pushforward from the normal distribution, for the purpose of offering a fair and
direct comparison and showcasing the advantage of our approach.

We would also like to point out that the loss~\eqref{eq:loss} satisfies
\[
	\mathcal{L}(\theta) = - \E_{\x\sim \pE}\log q_\theta(\x) = \mathrm{D}_{\text{KL}}
	\left(\pE(\cdot)\|q_\theta(\cdot) \right) -\E_{\x\sim \pE}\log \pE(\x).
\]
Therefore, one should \textit{not} expect the loss to approach zero during the
training process.

\subsection{Rosenbrock distribution}

\begin{figure}[!htbp]
	\centering
	\begin{subfigure}{0.45\textwidth}
		\centering
		\includegraphics[width=\textwidth]{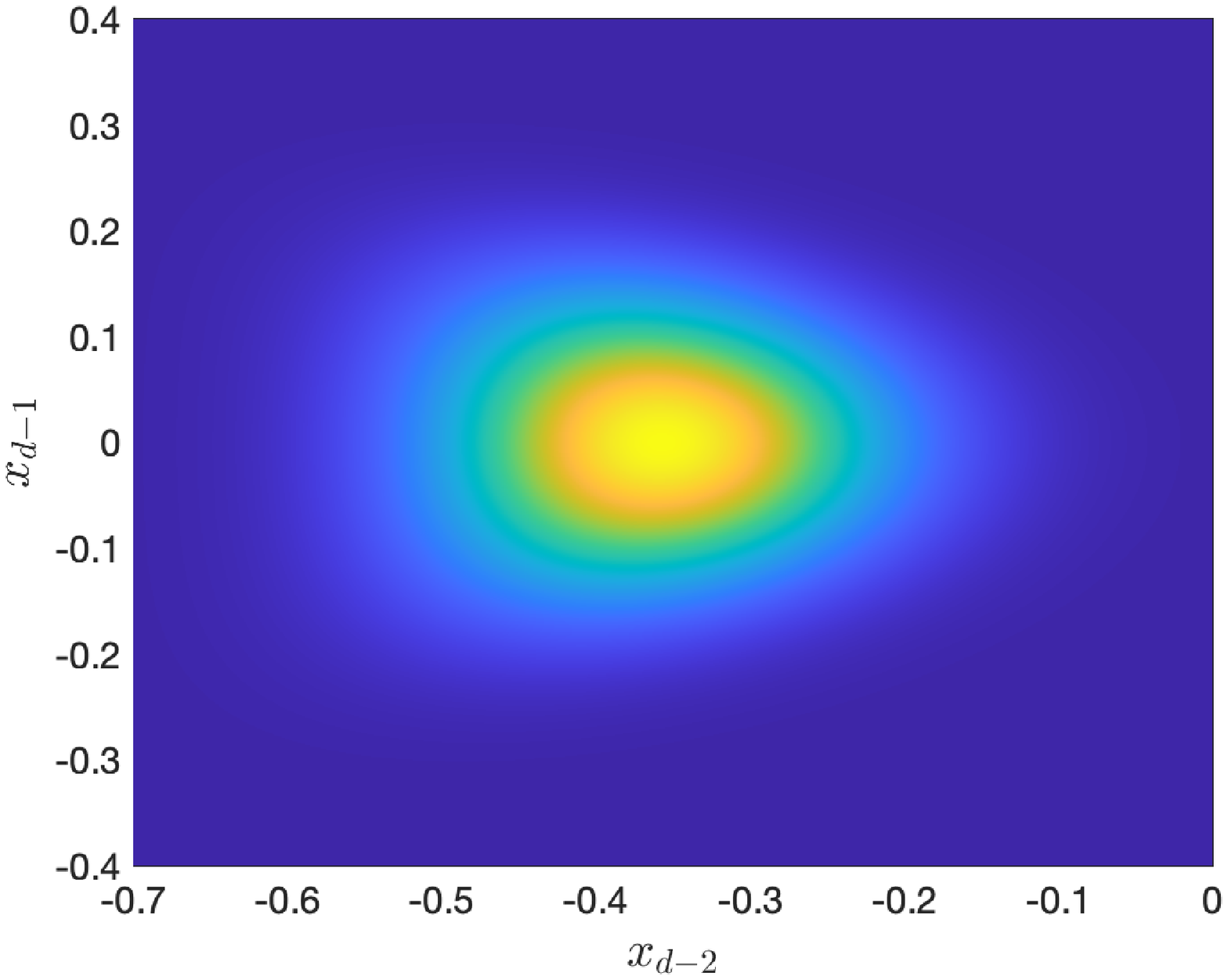}
		\caption{the $d-2$ and $d-1$-th dimensions}
	\end{subfigure}
	\hspace{1em}
	\begin{subfigure}{0.45\textwidth}
		\centering
		\includegraphics[width=\textwidth]{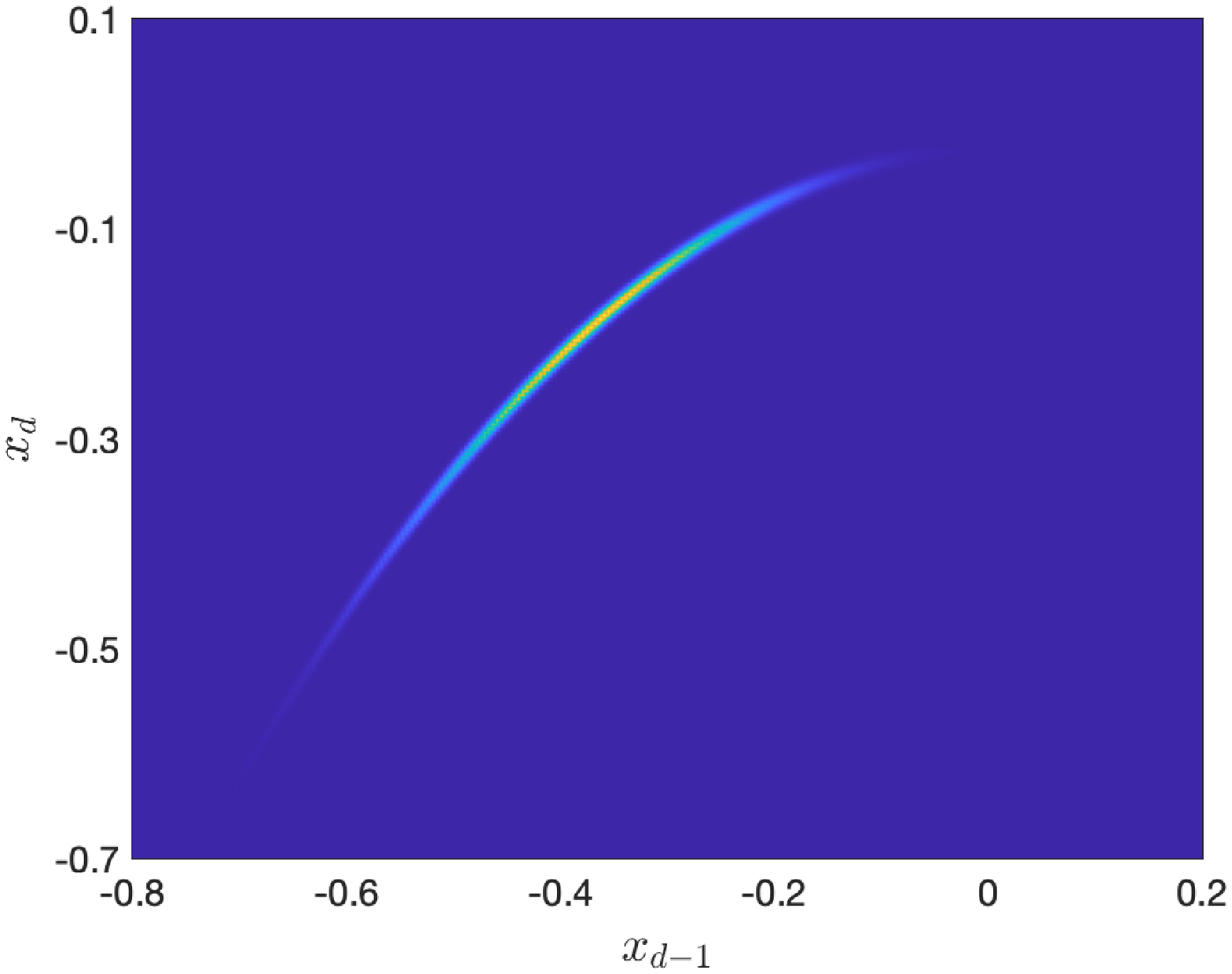}
		\caption{the $d-1$ and $d$-th dimensions}
	\end{subfigure}
	\caption{Marginal distributions of the Rosenbrock distribution: a singular structure appears on the last two dimensions.}
	\label{fig:rmargin}
\end{figure}

In this example, we consider the distribution induced by the Rosenbrock
function $v(x)$, \emph{i.e.} $p^*(\x)\propto \exp\left(-v(\x)/2\right)$, where
\[
	v(\x) = \sum_{i=1}^{d-1}\left[ c_i^2x_i^2+\left( c_{i+1}x_{i+1}+5(c_i^2x_i^2+1) \right)^2 \right].
\]
Here we set the dimension $d=10$, restrict all $x_i$ to the finite interval
$I=[-1,1]$, and select the scaling factor $c_i=2$ for $1\le i\le d-2$,
$c_{d-1}=7$, and $c_d = 200$, following the example used
by~\citet{dolgov2020approximation}. As shown in Figure~\ref{fig:rmargin},
although the Rosenbrock distribution is designed to be relatively isotropic in
the first $d-2$ variables, it is concentrated along a curve on the last two
dimensions. 

\begin{figure}[!htbp]
	\centering
	\begin{subfigure}{0.45\textwidth}
		\centering
		\includegraphics[width=\textwidth]{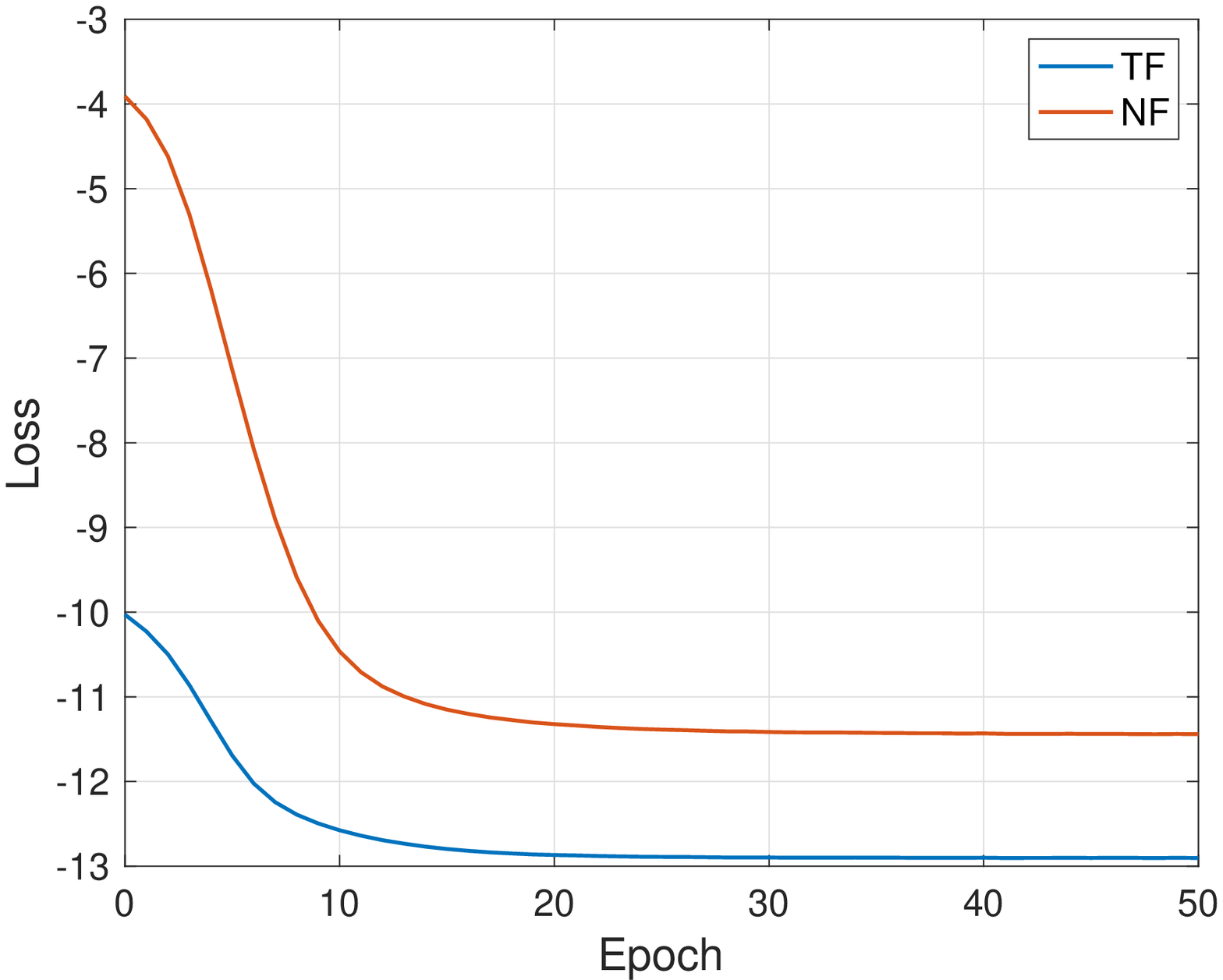}
		\caption{Training loss}
	\end{subfigure}
	\hspace{1em}
	\begin{subfigure}{0.45\textwidth}
		\centering
		\includegraphics[width=\textwidth]{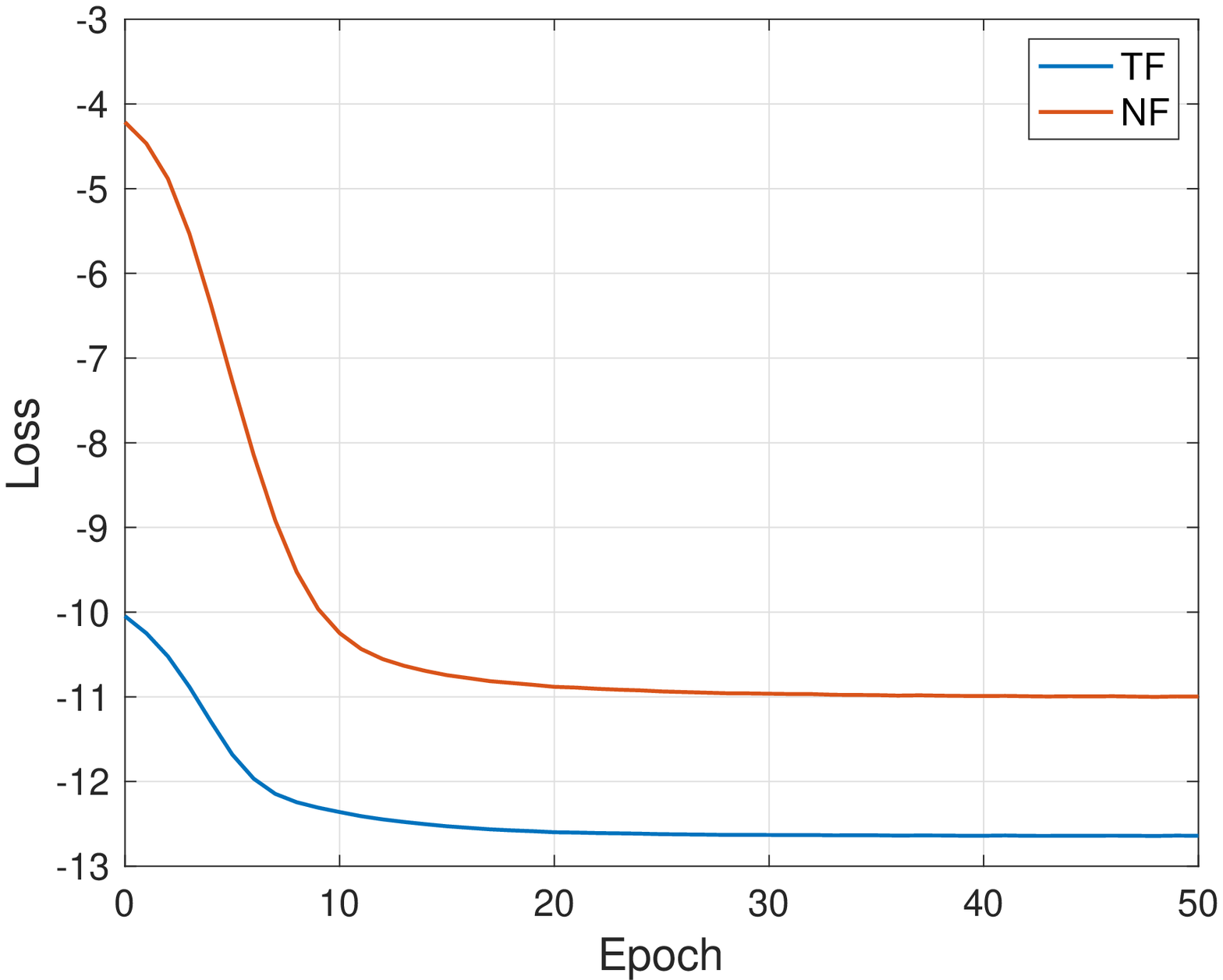}
		\caption{Test loss}
	\end{subfigure}
	\caption{Estimating the Rosenbrock distribution of dimension $d=10$ with
		sample size $N=10^5$: The initial and final loss of TF are both better
		than those of NF when training with the same neural network architecture
		and parameters. }
	\label{fig:rosenbrock}
\end{figure}

The experiment results are shown in Figure~\ref{fig:rosenbrock}. It can be seen
that our algorithm starts with a much lower loss compared to the normalizing
flow. This confirms our expectation that the approximate TT representation
$p^\TT(\x)$ serves as a much better base distribution $q(\x,0)$ than the normal
distribution in the flow model in terms of both initial and final losses.

\begin{figure}[!htbp]
	\centering
	\begin{subfigure}{0.32\textwidth}
		\centering
		\includegraphics[width=\textwidth]{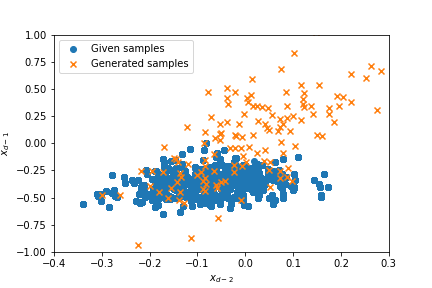}
		\caption{NF}
	\end{subfigure}
	\begin{subfigure}{0.32\textwidth}
		\centering
		\includegraphics[width=\textwidth]{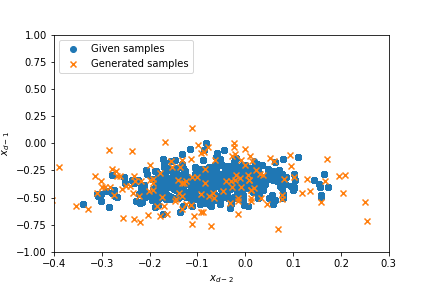}
		\caption{TT representation}
	\end{subfigure}
	\begin{subfigure}{0.32\textwidth}
		\centering
		\includegraphics[width=\textwidth]{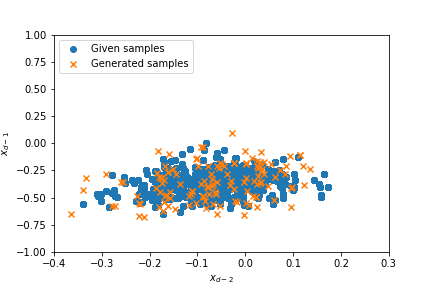}
		\caption{TF}
	\end{subfigure}
	\caption{Sampling results projected on the $d-2$ and $d-1$-th dimension for Rosenbrock distribution of dimension $d=10$: Samples from tensorizing flow agree better with the original distribution.}
	\label{fig:rsample32}
\end{figure}

\begin{figure}[!htbp]
	\centering
	\begin{subfigure}{0.32\textwidth}
		\centering
		\includegraphics[width=\textwidth]{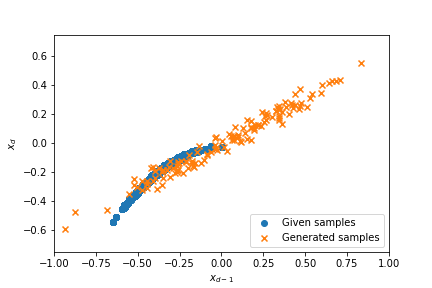}
		\caption{NF}
	\end{subfigure}
	\begin{subfigure}{0.32\textwidth}
		\centering
		\includegraphics[width=\textwidth]{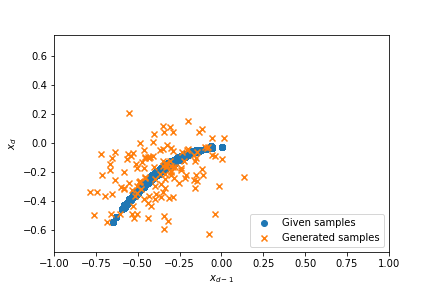}
		\caption{TT representation}
	\end{subfigure}
	\begin{subfigure}{0.32\textwidth}
		\centering
		\includegraphics[width=\textwidth]{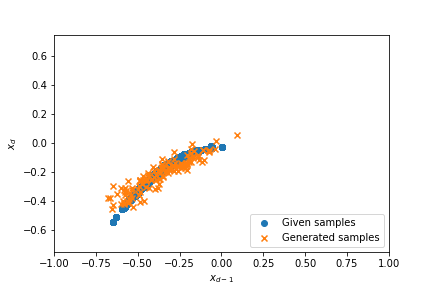}
		\caption{TF}
	\end{subfigure}
	\caption{Sampling results projected on the $d-1$ and $d$-th dimension for
		Rosenbrock distribution of dimension $d=10$: Tensor-train has
		limitations in representing the tail structure which in contrast can be
		learned satisfactorily by TF.}
	\label{fig:rsample21}
\end{figure}

Figure~\ref{fig:rsample32} and~\ref{fig:rsample21} illustrate several samples
generated by our method compared with the given samples. It is clear that
compared to using a normal distribution as the base distribution, using the
approximate TT representation better captures the rough structure of the target
distribution. In the work by~\citet{dolgov2020approximation}, extra fine grids
on the last two dimensions (4 and 32 times finer than the first $d-2$
dimensions, respectively) are adopted to deal with the singular tail structure.
However, since the empirical distribution is smoothed by kernel density
estimation in our approach, no other specific measures to the last two
dimensions are required for constructing the approximate TT representation and
the subsequent neural network-based flow corrects the estimation automatically.

\subsection{Ginzburg-Landau distribution}

The Ginzburg-Landau (GL) theory is a widely-used model for the study of phase
transition in statistical mechanics~\citep{hohenberg2015introduction}. The
general Ginzburg-Landau potential defined for a sufficiently smooth function
$x(\r):\Omega \rightarrow \R$, where $\Omega\subset\R^k$ is some domain with
suitable boundary conditions, is
\begin{equation}
	\mathcal{E}[x(\cdot)] = \int_\Omega\left[ \dfrac{\delta}{2}|\nabla_{\r} x(\r)|^2+\dfrac{1}{\delta}V(x(\r)) \right] \d \r,
	\label{eq:generalgl}
\end{equation}
where the potential $V(x)=\left(1-x^2\right)^2/4$. 

\subsubsection{1D Ginzburg-Landau distribution}

\begin{figure}[!htbp]
	\centering
	\begin{subfigure}{0.38\textwidth}
		\centering
		\includegraphics[width=\textwidth]{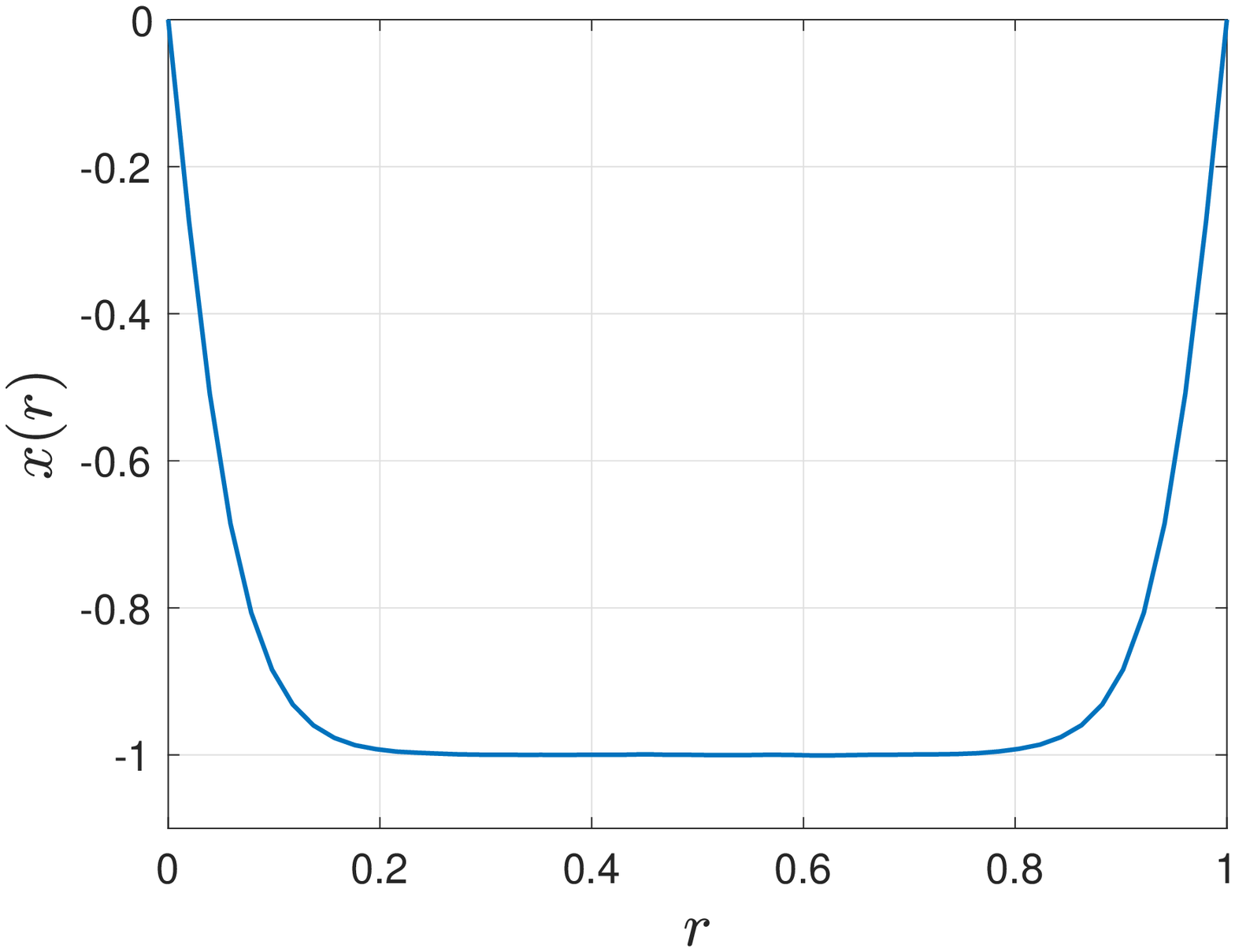}
		\caption{$x_-$}
	\end{subfigure}
	\hspace{1em}
	\begin{subfigure}{0.38\textwidth}
		\centering
		\includegraphics[width=\textwidth]{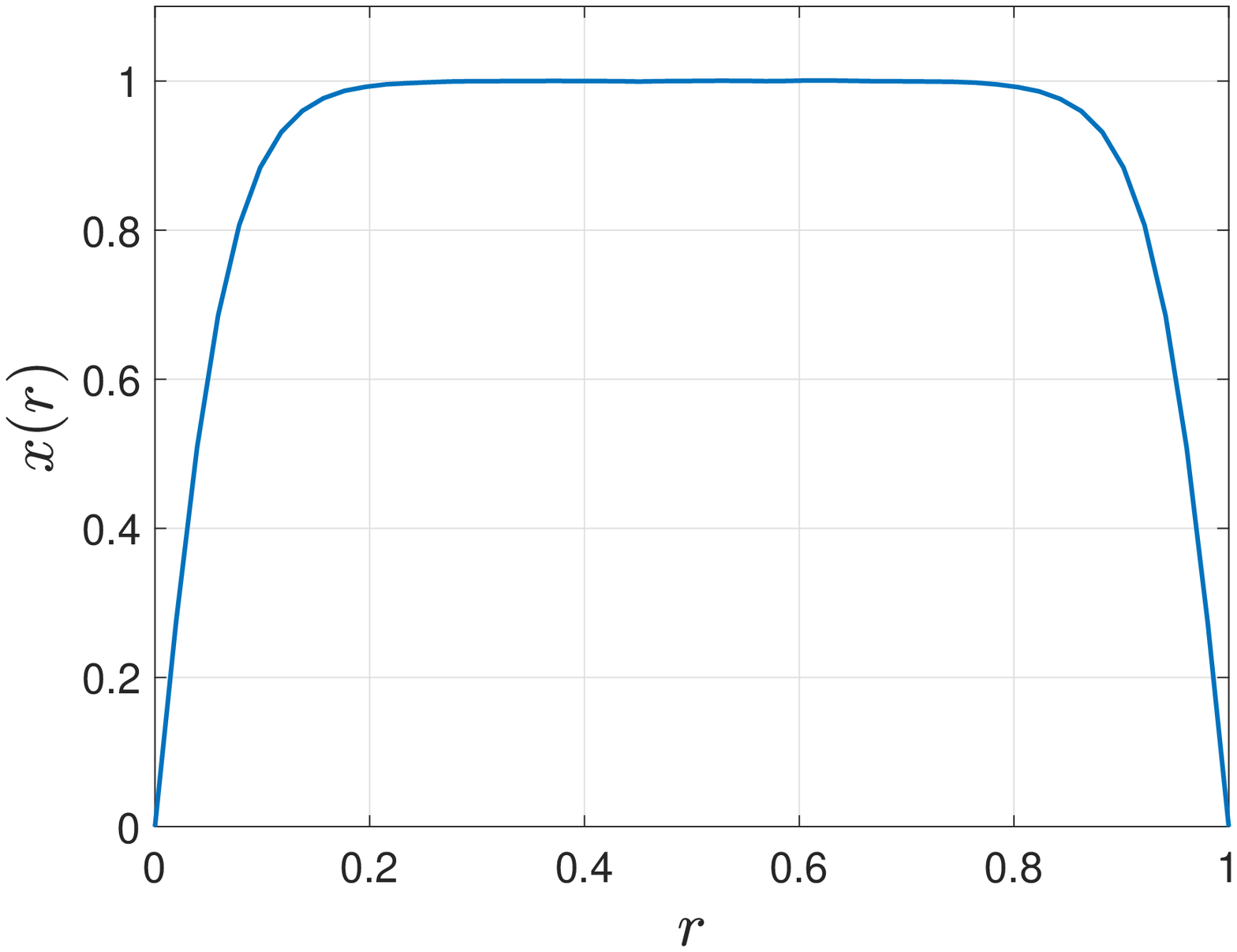}
		\caption{$x_+$}
	\end{subfigure}
	\caption{Two local minimizers of the 1D GL potential with $\delta=0.05$ and
	$L=1$. }
	\label{fig:minimizers}
\end{figure}

In the 1-dimensional case of the Ginzburg-Landau potential, we fix the domain
$\Omega=[0,L]$ and discretize the function $x(\r)$ with the vector $\x =
(x_0,\ldots,x_{d+1})$ consisting of its values on the uniform grid
$(ih)_{i=0}^{d+1}$ with Dirichlet boundary condition $x_0 = x_{d+1} = 0$ and
grid size $h = {L}/(d+1)$. Consequently, by using the first-order finite
difference scheme and estimating the integral in~\eqref{eq:generalgl} by the
right Riemann sum, the 1D Ginzburg-Landau potential is approximated by
\begin{equation}
	E(\x) = \sum_{i=1}^{d+1}\left[\dfrac{\delta}{2}\left(\dfrac{x_i-x_{i-1}}{h}\right)^2+\dfrac{1}{4\delta}\left(1-x_i^2\right)^2\right]
	\label{eq:gl1d}
\end{equation}
and its associated Boltzmann distribution satisfies $p^*(\x)\propto \exp(-\beta
E(x))$, where $\beta$ is the inverse temperature.  As mentioned
by~\citet{weinan2004minimum}, most of the states $\x$ of interest lie within the
range between $\x_-$ and $\x_+$, the two minimizers of the 1D Ginzburg-Landau
potential~\eqref{eq:gl1d} shown in Figure~\ref{fig:minimizers}. Thus we choose
$I=[-3,3]$ as the range for each $x_i$ in the discretization $\x$.

\begin{figure}[!htbp]
	\centering
	\begin{subfigure}{0.45\textwidth}
		\centering
		\includegraphics[width=\textwidth]{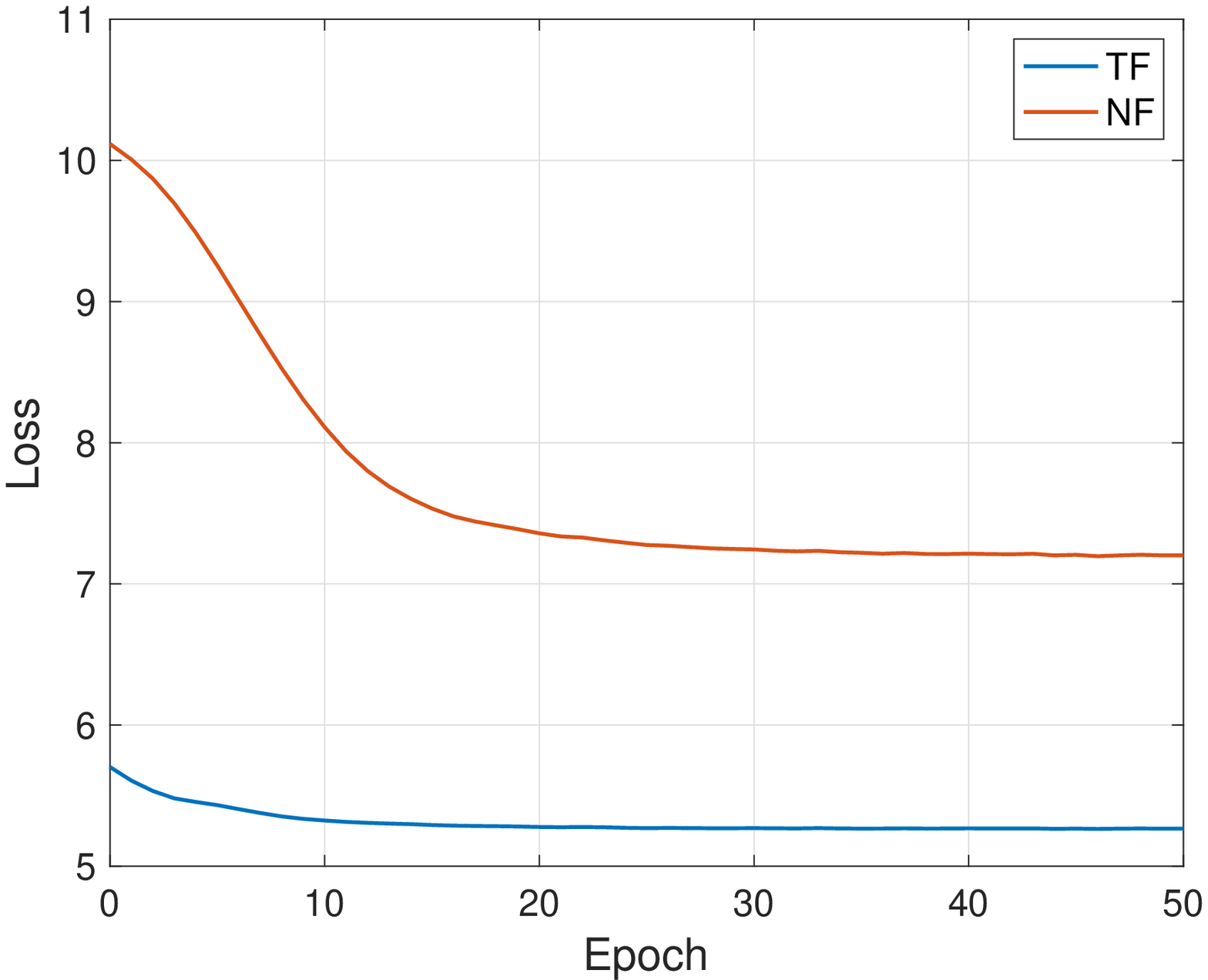}
		\caption{Training loss}
	\end{subfigure}
	\hspace{1em}
	\begin{subfigure}{0.45\textwidth}
		\centering
		\includegraphics[width=\textwidth]{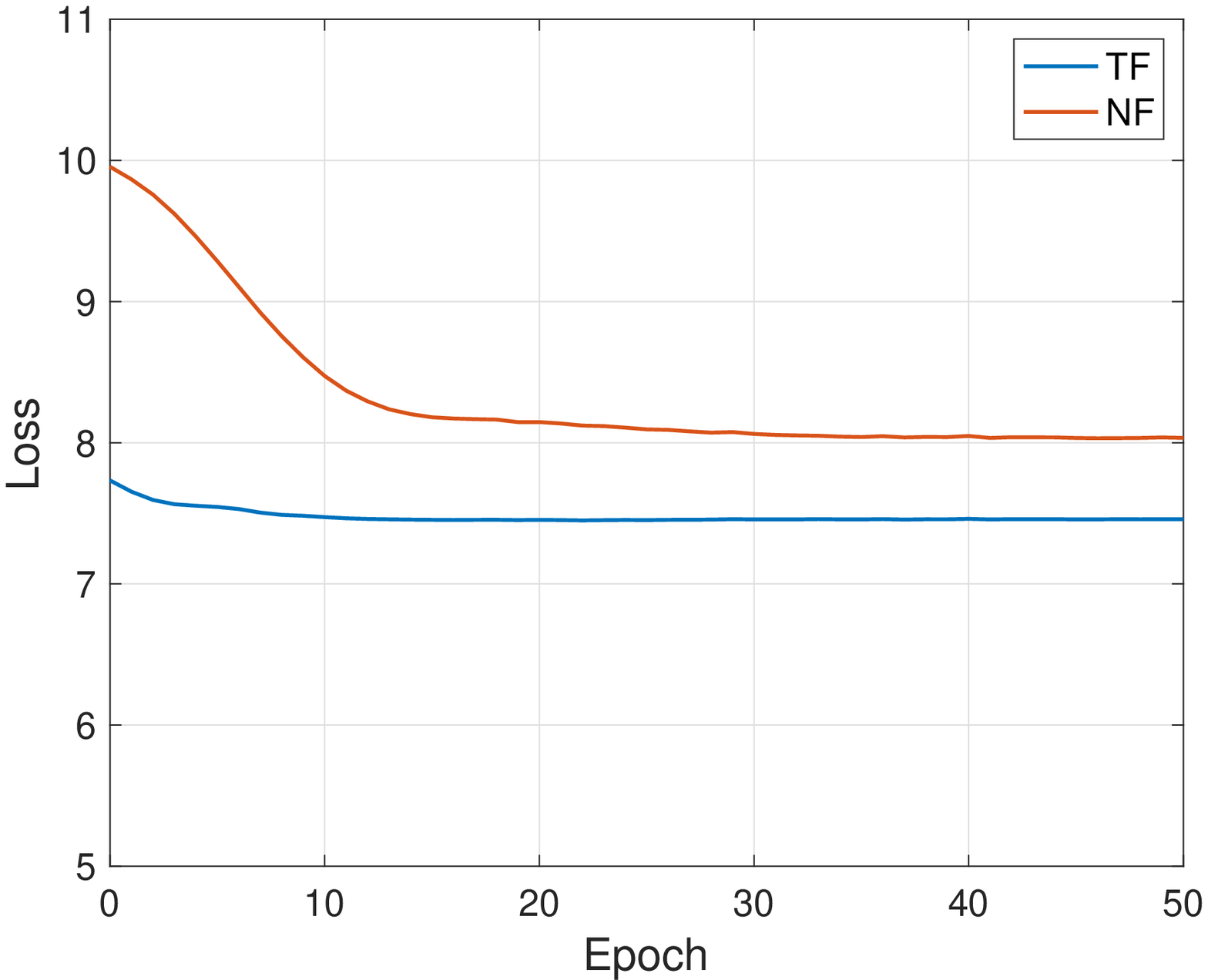}
		\caption{Test loss}
	\end{subfigure}
	\caption{Estimating the 1D GL distribution of dimension $d=8$ with sample
	size $N=10^4$. }
	\label{fig:gl1d}
\end{figure}

\begin{figure}[!htbp]
	\centering
	\begin{subfigure}{0.45\textwidth}
		\centering
		\includegraphics[width=\textwidth]{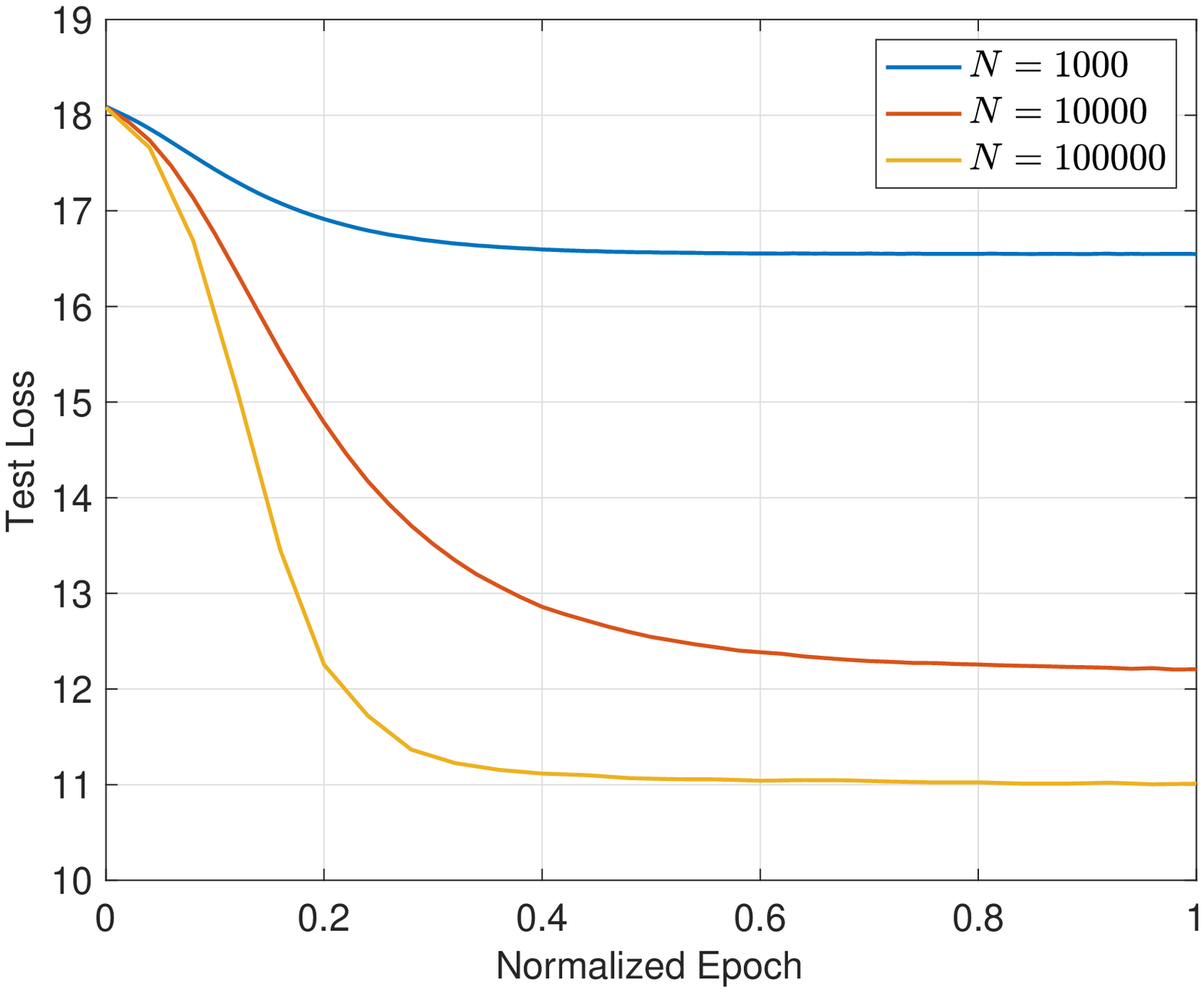}
		\caption{Normalizing flow}
	\end{subfigure}
	\hspace{1em}
	\begin{subfigure}{0.45\textwidth}
		\centering
		\includegraphics[width=\textwidth]{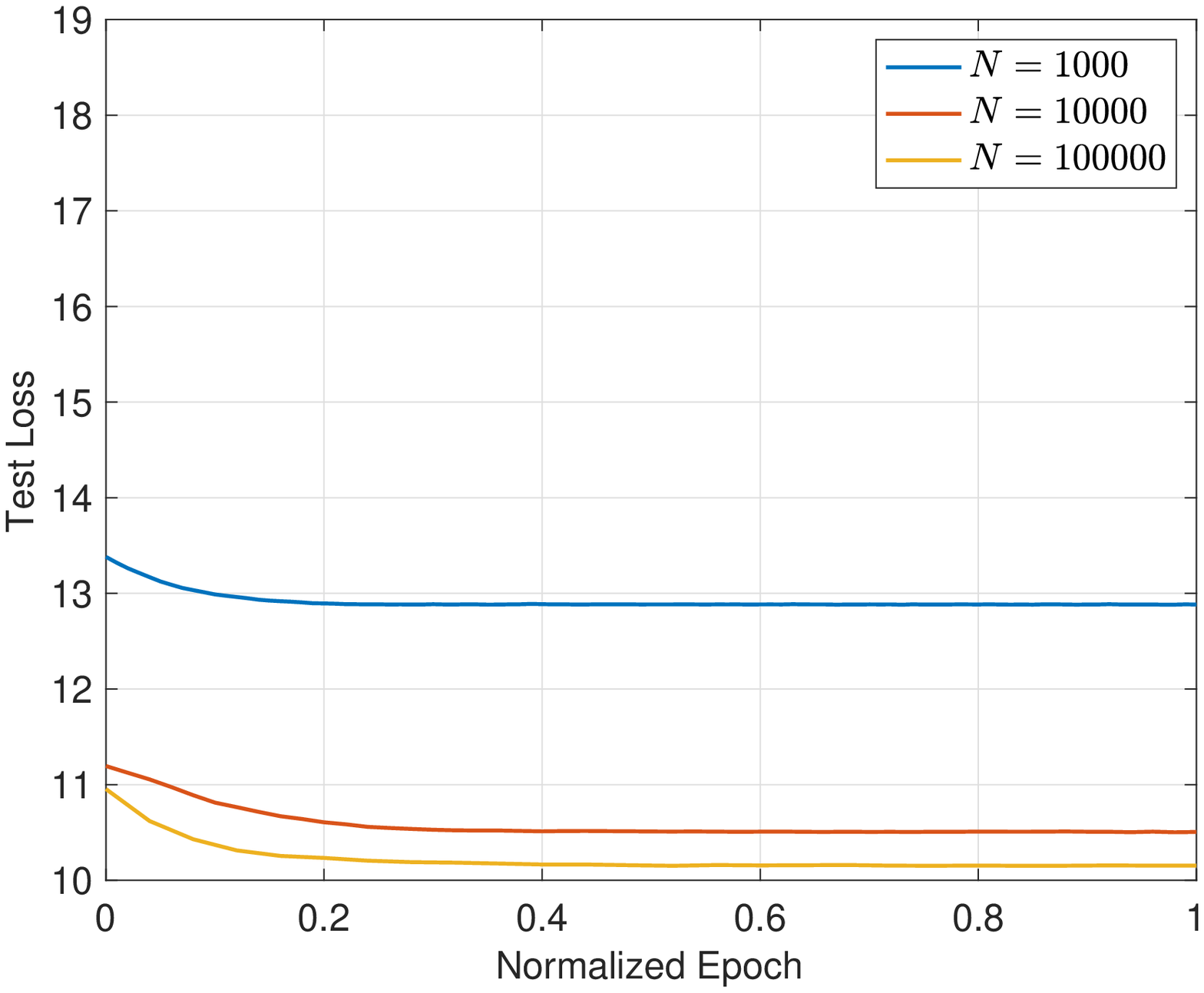}
		\caption{Tensorizing flow}
	\end{subfigure}
	\caption{Comparison of test loss for estimating 1D GL distribution of
		dimension $d=16$ with different sample sizes $N$: TF yields better
		results with much fewer samples than NF of the same neural network
		architecture.    
	}
	\label{fig:samplesize}
\end{figure}

\begin{figure}[!htbp]
	\centering
	\begin{subfigure}{0.45\textwidth}
		\centering
		\includegraphics[width=\textwidth]{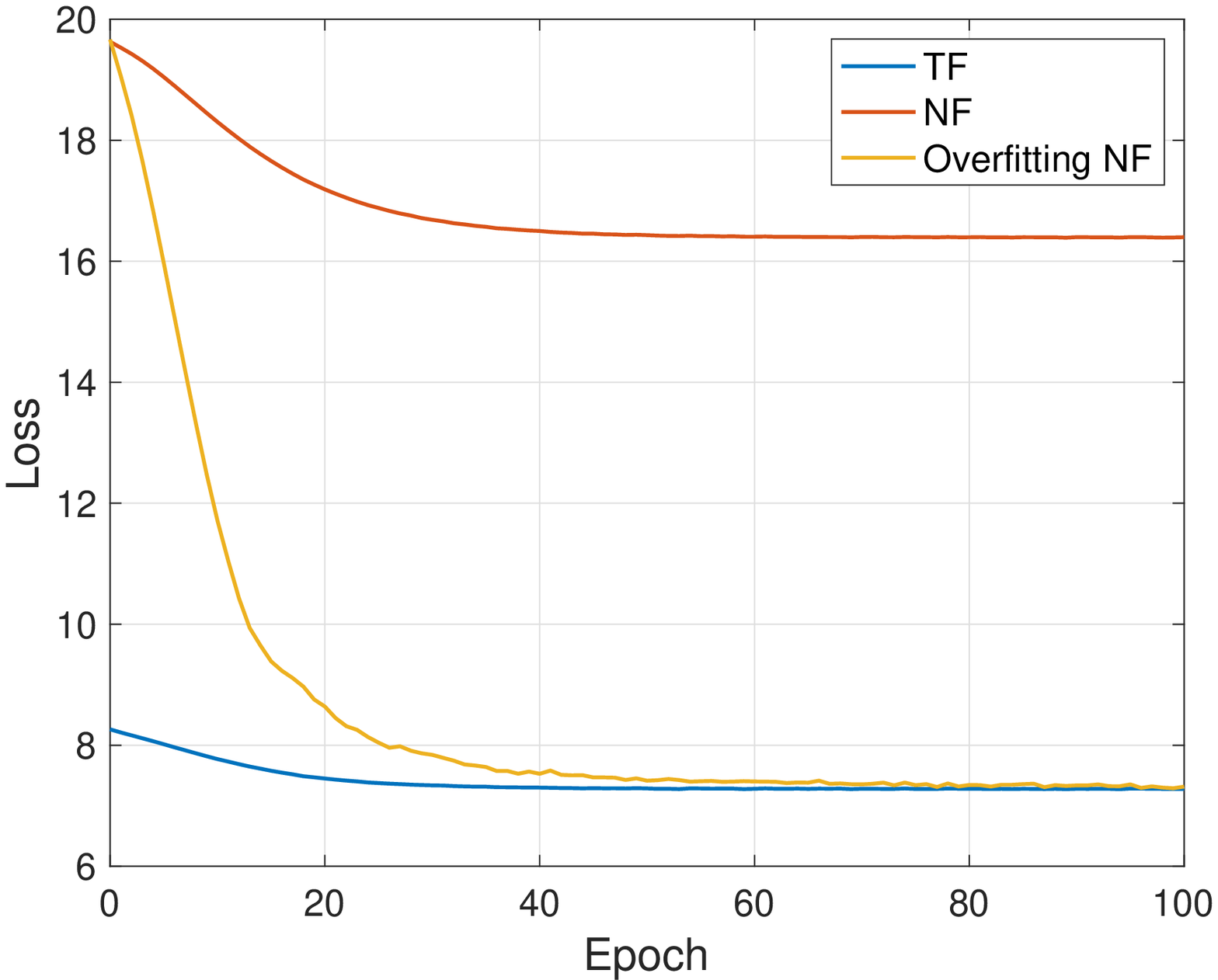}
		\caption{Training loss}
		\label{fig:gl1d2a}
	\end{subfigure}
	\hspace{1em}
	\begin{subfigure}{0.45\textwidth}
		\centering
		\includegraphics[width=\textwidth]{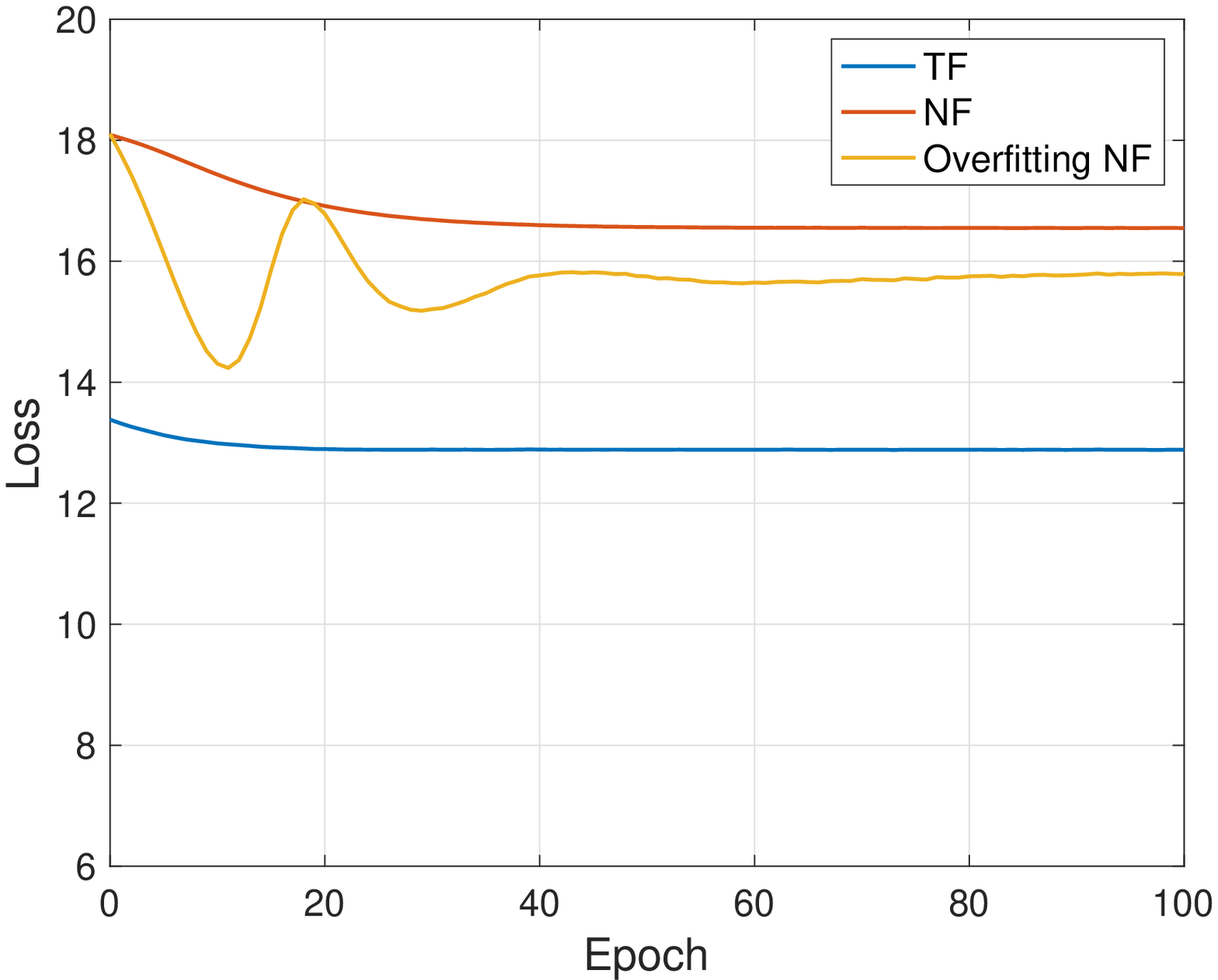}
		\caption{Test loss}
		\label{fig:gl1d2b}
	\end{subfigure}
	\caption{Estimating the 1D GL distribution of dimension $d=16$ with sample
		size $N=1000$: NF with $10^6$ parameters overfits significantly compared
		with TF with $10^4$ parameters. }
	\label{fig:gl1d2}
\end{figure}

The results for the case where $d =8$, $\beta = 3$, $\delta = 0.5$, and $h=1$
are shown in Figure~\ref{fig:gl1d}. We can see that as in the previous example,
the loss of the tensorizing flow starts lower than normalizing flow, and ends up
better than its counterpart.

In order to demonstrate the effect of the sample size on our method, we also
perform experiments to the setting $d = 16$, $\beta = 3$, $\delta = 1$, and
$h=1$ with sample sizes $10^3$, $10^4$, and $10^5$. The training parameters
including the number of epochs, learning rate, \textit{etc.}, are adjusted
proportionally to offer a fair comparison between these experiments (see
Table~\ref{tab:hyper}). The results are shown in Figure~\ref{fig:samplesize},
where the horizontal axis is the normalized epoch, \emph{i.e.} epoch divided by
the total number of epochs, and the vertical axis is the test loss computed
using a common test set of size $5\times 10^3$. As the performance of the
normalizing flow keeps improving when the sample size increases, a larger sample
set also yields a better TT representation $p^\TT$ to start with and eventually
a better density estimation $p^\TF$ after training. Also,
Figure~\ref{fig:samplesize} indicates that our method produces a better density
estimation with $10^4$ samples than that with $10^5$ samples by normalizing
flow, which corroborates the efficiency of our method in the sense of samples
required for reaching certain accuracy of the density estimation.

In Figure~\ref{fig:gl1d2}, we present another example with $d =16$, $\beta = 3$,
$\delta = 1$, $h=1$, and $N=10^3$ trying to understand why normalizing flow
cannot achieve the same test loss as tensorizing flow. With the same
architecture, indeed normalizing flow has higher training and test losses than
tensorizing flow (see the red curves in Figure~\ref{fig:gl1d2}). In order to
improve the training loss of normalizing flow, we use a relatively
overparameterized NN for normalizing flow (see the corresponding parameters in
Table~\ref{tab:hyper}) so that its training loss matches with that of the
tensorizing flow (see the yellow curve in Figure~\ref{fig:gl1d2a}). It is clear
that with a matching training loss, normalizing flow overfits significantly (see
the yellow curve in Figure~\ref{fig:gl1d2b}), demonstrating that tensorizing
flow provides much better generalization and is not prone to overfit, since it
only uses a relatively small and less expressive neural network.

\subsubsection{2D Ginzburg-Landau distribution}

\begin{figure}[!htbp]
	\centering
	\begin{tikzpicture}
		\node[draw, circle, scale = 0.8](00) at (0,0) {$x_{1,1}$}; \node[draw,
		circle, scale = 0.8](10) at (1.2,0) {$x_{2,1}$}; \node[draw, circle,
		scale = 0.8](20) at (2.4,0) {$x_{3,1}$}; \node[draw, circle, scale =
		0.8](30) at (3.6,0) {$x_{4,1}$};
		        
		\node[draw, circle, scale = 0.8](01) at (0,1.2) {$x_{1,2}$}; \node[draw,
		circle, scale = 0.8](11) at (1.2,1.2) {$x_{2,2}$}; \node[draw, circle,
		scale = 0.8](21) at (2.4,1.2) {$x_{3,2}$}; \node[draw, circle, scale =
		0.8](31) at (3.6,1.2) {$x_{4,2}$};
		        
		\node[draw, circle, scale = 0.8](02) at (0,2.4) {$x_{1,3}$}; \node[draw,
		circle, scale = 0.8](12) at (1.2,2.4) {$x_{2,3}$}; \node[draw, circle,
		scale = 0.8](22) at (2.4,2.4) {$x_{3,3}$}; \node[draw, circle, scale =
		0.8](32) at (3.6,2.4) {$x_{4,3}$};
		        
		\node[draw, circle, scale = 0.8](03) at (0,3.6) {$x_{1,4}$}; \node[draw,
		circle, scale = 0.8](13) at (1.2,3.6) {$x_{2,4}$}; \node[draw, circle,
		scale = 0.8](23) at (2.4,3.6) {$x_{3,4}$}; \node[draw, circle, scale =
		0.8](33) at (3.6,3.6) {$x_{4,4}$};
		        
		\draw (00) -- (10) -- (20) -- (30) to[out = 20, in = 160] (00); \draw
		(01) -- (11) -- (21) -- (31) to[out = 20, in = 160] (01); \draw (02) --
		(12) -- (22) -- (32) to[out = 20, in = 160] (02); \draw (03) -- (13) --
		(23) -- (33) to[out = 20, in = 160] (03);
		        
		\draw (00) -- (01) -- (02) -- (03) to[out = 70, in = -70] (00); \draw
		(10) -- (11) -- (12) -- (13) to[out = 70, in = -70] (10); \draw (20) --
		(21) -- (22) -- (23) to[out = 70, in = -70] (20); \draw (30) -- (31) --
		(32) -- (33) to[out = 70, in = -70] (30);
		        
		\draw[->,red,thick] (00) -- (01); \draw[->,red,thick] (01) -- (02);
		\draw[->,red,thick] (02) -- (03); \draw[->,red,thick] (03) -- (13);
		\draw[->,red,thick] (13) -- (12); \draw[->,red,thick] (12) -- (11);
		\draw[->,red,thick] (11) -- (10); \draw[->,red,thick] (10) -- (20);
		\draw[->,red,thick] (20) -- (21); \draw[->,red,thick] (21) -- (22);
		\draw[->,red,thick] (22) -- (23); \draw[->,red,thick] (23) -- (33);
		\draw[->,red,thick] (33) -- (32); \draw[->,red,thick] (32) -- (31);
		\draw[->,red,thick] (31) -- (30);
	\end{tikzpicture}
	\caption{Periodic boundary condition of the 2D Ginzburg-Landau distribution and the ``snake ordering'' of the matrix variable $\x$ of dimension $d=4\times4$:  pairs of variables with energy functions between them are connected by black lines and the vectorization of $\x$ is along the order indicated by the red arrow path.}
	\label{fig:pbc}
\end{figure}
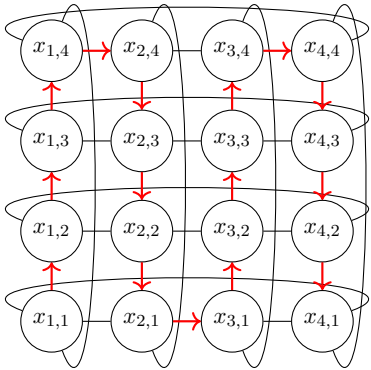

In the 2-dimensional case of the Ginzburg-Landau potential, we fix the domain
$\Omega=[0,L]^2$ and discretize the function $x(\r)$ by the matrix $\x =
\left(x_{i,j}\right)_{i,j=1}^{\sqrt{d}}$ where $x_{i,j}$ represents its value at
the grid point $((i-1)h,(j-1)h)$ with grid size $h = {L}/(\sqrt{d}-1)$. With a
similar discretization procedure as in the 1-dimensional case, the probability
density function of the 2D Ginzburg-Landau distribution satisfies $p^*(\x)\propto
\exp(-\beta E(\x))$, where
\begin{equation}
	E(\x)=
	\sum_{i=1}^{\sqrt{d}}\sum_{j=1}^{\sqrt{d}}\left[\dfrac{\delta}{2}\left(\left(\dfrac{x_{i,j}-x_{i-1,j}}{h}\right)^2+\left(\dfrac{x_{i,j}-x_{i,j-1}}{h}\right)^2\right)+\dfrac{1}{4\delta}\left(1-x_{i,j}^2\right)^2\right]
	\label{eq:gl2d}
\end{equation}
and the periodic boundary condition is adopted, \emph{i.e.} $x_{0,j} =
x_{\sqrt{d},j}$ for $1\le j\le \sqrt{d}$ and $x_{i,0} = x_{i,\sqrt{d}}$ for
$1\le i\le \sqrt{d}$, as shown in Figure~\ref{fig:pbc}.

Unlike the 1D Ginzburg-Landau model, the 2D Ginzburg-Landau is not Markovian. In
order to obtain an approximate TT representation using the algorithm proposed in
Section~\ref{sec:algstpA}, we adopt the ``snake ordering'' when vectorizing the
matrix $\x$ in the order demonstrated by the red arrow path in
Figure~\ref{fig:pbc} to exploit the Markovian structure of~\eqref{eq:gl2d} to
the largest extent.

\begin{figure}[!htbp]
	\centering
	\begin{subfigure}{0.45\textwidth}
		\centering
		\includegraphics[width=\textwidth]{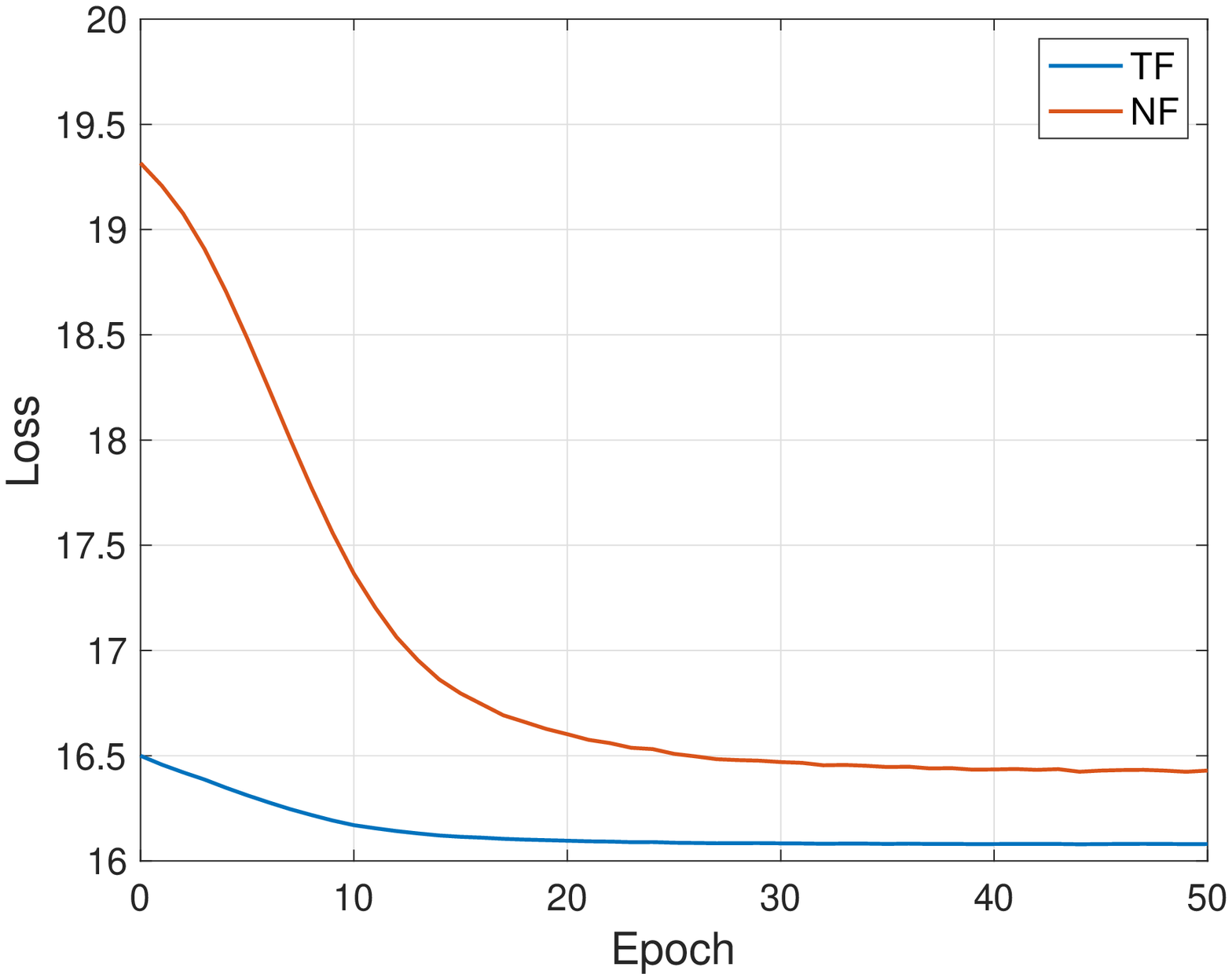}
		\caption{Training loss}
	\end{subfigure}
	\hspace{1em}
	\begin{subfigure}{0.45\textwidth}
		\centering
		\includegraphics[width=\textwidth]{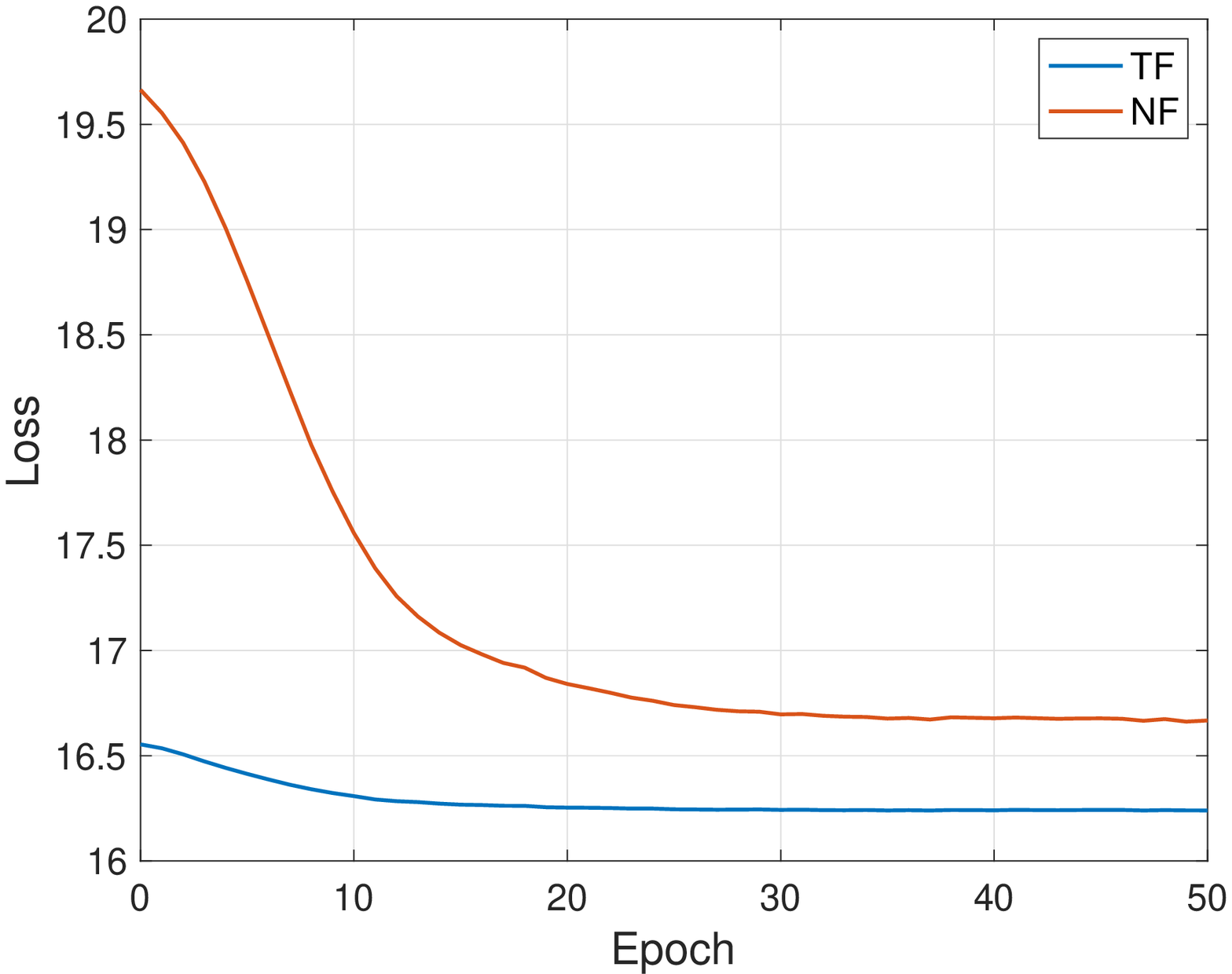}
		\caption{Test loss}
	\end{subfigure}
	\caption{Estimating the 2D Ginzburg-Landau distribution of dimension
		$d=4\times4$ with sample size $N=10^4$: TF learns a complicated
		non-Markovian density. }
	\label{fig:gl2d}
\end{figure}

In our example, we set the dimension $d=4\times4$, $\beta = 1.5$, $\delta = 1$,
$h = 1$. The range of each $x_{i,j}$ is also assumed to be within $I=[-3,3]$.
The experiment results of this example are shown in Figure~\ref{fig:gl2d}, which
further confirms the effectiveness of our algorithm over either normalizing flow
or the TT representation when dealing with more complicated non-Markovian
distributions.

\section{Discussions}
\label{sec:conclusions}

We proposed a generative model for high-dimensional density estimation from a
finite collection of samples. Using a sketching technique, we construct an
approximate tensor-train representation efficiently. When constructing the
tensor-train, we adopt kernel density estimation for estimating the required
low-dimensional marginals. Starting from the tensor-train representation as the
base distribution, we perform the continuous-time flow model to further refine
our density estimation, featuring a potential function parameterized by a neural
network and fast calculation of both the forward and inverse map by Runge-Kutta
scheme.

Several experiments demonstrate that our method evidently outperforms
normalizing flow of similar architectures and is capable of dealing with
distributions of certain singularity as well as non-Markovian models, for which
traditional tensor-train methods may run into difficulties. Due to the
near-identity nature of the tensorizing flow, a relatively simple neural network
suffices for the flow model, which is much easier to train and less prone to
overfitting compared with normalizing flow.

However, our method is still confined to the presumed Markovian structure of the
distribution and future research may focus on designing a more adaptive scheme
for non-Markovian models with more complicated graph structure. Furthermore,
although the Legendre polynomials are used as the expansion basis in this work,
our method is open to other expansion basis, including the Chebyshev polynomials
and the Fourier basis. 


\acks{Yinuo Ren and Lexing Ying are partially supported by National Science Foundation under Award No. DMS-2011699. Hongli Zhao and Yuehaw Khoo are partially supported by National Science Foundation under Award No. DMS-2111563. Yuehaw Khoo is partially supported by U.S. Department of Energy, Office of Science under Award No. DE-SC0022232.}


\newpage

\appendix
\section{Proof of Proposition~\ref{prop:cde}}
\label{app:proof}

In this appendix we prove the Proposition~\ref{prop:cde} from
Section~\ref{sec:algstpA}:

\noindent
{\bf Proof of Proposition~\ref{prop:cde}.} For $2\le k\le d$, it suffices for us
to consider the $k$-th equation in~\eqref{eq:cde}:
\begin{equation}
	\sum_{\alpha_{k-1}=1}^{r_{k-1}}\Phi_{k-1}(x_{1:k-1};\alpha_{k-1})G_k(\alpha_{k-1};x_k,\alpha_k)=\Phi_k(x_{1:k-1};x_k,\alpha_k).
	\label{eq:kth}
\end{equation}
    
By Definition~\ref{def:finiterank}, there exist orthonormal right singular
vectors $$\{\Psi_{k-1}(\alpha_{k-1};x_{k:d})\}_{1\le \alpha_{k-1} \le r_{k-1}}
\subset L^2(I^{d-k+1})$$ of $p(x_{1:k-1};x_{k:d})$ and
$$\{\Psi_k(\alpha_{k};x_{k+1:d})\}_{1\le \alpha_k\le r_{k}} \subset
L^2(I^{d-k})$$ of $p(x_{1:k};x_{k+1:d})$, and corresponding singular values
$\sigma_{k-1}(1)\geq \cdots\geq \sigma_{k-1}(r_{k-1})$ and $\sigma_{k}(1)\geq
\cdots\geq \sigma_{k}(r_{k})$, satisfying 
\begin{equation}
	p(x_{1:k-1};x_{k:d})= \sum_{\alpha_{k-1}=1}^{r_{k-1}}\sigma_{k-1}(\alpha_{k-1})\Phi_{k-1}(x_{1:k-1};\alpha_{k-1})\Psi_{k-1}(\alpha_{k-1};x_{k:d}),
	\label{eq:k-1}
\end{equation}
and
\[
	p(x_{1:k};x_{k+1:d})= \sum_{\alpha_{k}=1}^{r_{k}}\sigma_{k}(\alpha_{k})\Phi_{k}(x_{1:k};\alpha_{k})\Psi_k(\alpha_{k};x_{k+1:d}).
\]
Define
$\Xi_k(x_{k+1:d};\alpha_k)=\sigma_k(\alpha_k)^{-1}\Psi_k(\alpha_k;x_{k+1:d})$.
It is easy to check that
\begin{equation*}
	\begin{aligned}
		  & \int_{I^{d-k}} p(x_{1:k};x_{k+1:d})\Xi_k(x_{k+1:d};\alpha_k) \d x_{k+1:d}                                                                                                                     \\
		= & \int_{I^{d-k}} \sum_{\alpha'_k=1}^{r_{k}}\sigma_{k}(\alpha'_{k})\sigma_{k}(\alpha_{k})^{-1}\Phi_{k}(x_{1:k};\alpha'_{k})\Psi_k(\alpha'_{k};x_{k+1:d})\Psi_k(\alpha_{k};x_{k+1:d})\d x_{k+1:d} \\
		= & \Phi_{k}(x_{1:k};\alpha_{k}).                                                                                                                                                                 
	\end{aligned}
\end{equation*}
Therefore, by contracting $\Xi_k(x_{k+1:d};\alpha_k)$ to both sides
of~\eqref{eq:k-1}, we have
\begin{equation*}
	\begin{aligned}
		  & \Phi_{k}(x_{1:k};\alpha_{k}) = \int_{I^{d-k}} p(x_{1:k-1};x_{k:d}) \Xi_k(x_{k+1:d};\alpha_k)\d x_{k+1:d}                                                                         \\
		= & \sum_{\alpha_{k-1}=1}^{r_{k-1}}\sigma_{k-1}(\alpha_{k-1})\Phi_{k-1}(x_{1:k-1};\alpha_{k-1})\int_{I^{d-k}} \Psi_{k-1}(\alpha_{k-1};x_{k:d})\Xi_k(x_{k+1:d};\alpha_k)\d x_{k+1:d}, 
	\end{aligned}
\end{equation*}
and consequently
$$
G_k(\alpha_{k-1};x_k,\alpha_k)=\sigma_{k-1}(\alpha_{k-1})\int_{I^{d-k}} \Psi_{k-1}(\alpha_{k-1};x_{k:d})\Xi_k(x_{k+1:d};\alpha_k)\d x_{k+1:d}
$$
solves the equation~\eqref{eq:kth}.
    
The uniqueness of the solution is guaranteed by the orthogonality of the
functions $\{\Psi_{k-1}(\alpha_{k-1};x_{k:d})\}_{1\le \alpha_{k-1} \le r_{k-1}}$
by definition. Once $G_k$ are ready, it is easy to check the validity
of~\eqref{eq:lowranktt} by plugging the CDE in~\eqref{eq:cde} one into the next
successively.

\section{Hyperparameters}
\label{app:hyper}

In this section, we present the hyperparameters of our tensorizing flow
algorithm used for each examples in Section~\ref{sec:experiments}. For
simplicity, we choose the internal ranks $r_k=2$ for $1\le k\le d-1$, and the
number of quadrature points $l=20$ for all numerical integrations involved. We
set the time horizon $T=0.2$ with stepsize $\tau=0.01$ in the flow model. We
generate ${N}/{2}$ samples separately from the training samples as the test
samples. The rest of hyperparameters are organized in Table~\ref{tab:hyper}.

\begin{table}[!htbp]
	\small
	\begin{center}
		\begin{tabular}{cccccccccc} 
			\toprule
			\bf Example                                              & \bf
			Instance               & $N$  & $M$ & $N_{\text{batch}}$ & $D$  & LR
			& WD   & $\gamma$ \\ 
			\midrule
			\bf Rosenbrock(Figure~\ref{fig:rosenbrock})              & \bf TF/NF
			& 1e+5 & 30  & 5e+3               & 64   & 5e-4 & 2e-3 & 0.9      \\
			\midrule
			\bf 1D GL(Figure~\ref{fig:gl1d})                         & \bf TF/NF
			& 1e+4 & 25  & 5e+3               & 128  & 5e-3 & 1e-3 & 0.9      \\
			\midrule
			\multirow{3}{*}{\bf 1D GL(Figure~\ref{fig:samplesize}) } &
			\multirow{3}{*}{\bf TF/NF} & 1e+3 & 25  & 1e+3               & 128
			& 5e-3 & 1e-3 & 0.9      \\
			                                                         &
			                                                         & 1e+4 & 25
			                                                         & 5e+3
			                                                         & 128  &
			                                                         5e-3 & 1e-3
			                                                         & 0.9
			                                                         \\
			                                                         &
			                                                         & 1e+5 & 25
			                                                         & 5e+3
			                                                         & 128  &
			                                                         2e-3 & 1e-3
			                                                         & 0.85
			                                                         \\
			\midrule
			\multirow{2}{*}{\bf 1D GL(Figure~\ref{fig:gl1d2}) }      & \bf TF/NF
			& 1e+3 & 25  & 1e+3               & 128  & 5e-3 & 1e-3 & 0.9      \\
			                                                         & \bf
			                                                         Overfitting
			                                                         NF
			                                                         & 1e+3 & 25
			                                                         & 1e+3
			                                                         & 1024 &
			                                                         2e-3 & 1e-3
			                                                         & 0.9
			                                                         \\
			\midrule
			\bf 2D GL(Figure~\ref{fig:gl2d})                         & \bf TF/NF
			& 1e+4 & 25  & 5e+3               & 128  & 5e-3 & 1e-3 & 0.9      \\
			\bottomrule
		\end{tabular}
	\end{center}
	\caption{Hyperparameters used in the examples}  
	\label{tab:hyper}
\end{table}

\vskip 0.2in
\bibliography{sample}

\end{document}